\documentclass{article}

\PassOptionsToPackage{numbers}{natbib}

\usepackage[preprint]{Neurips_2025_Style_Guide/neurips_2025}

\usepackage[utf8]{inputenc} 
\usepackage[T1]{fontenc}    
\usepackage{hyperref}       
\usepackage{url}            
\usepackage{booktabs}       
\usepackage{amsfonts}       
\usepackage{nicefrac}       
\usepackage{microtype}      
\usepackage{xcolor}         
\usepackage{multirow}
\usepackage{pgfplots}       
\pgfplotsset{compat=1.18}
\usepackage{subcaption}

\title{Inference Scaled GraphRAG: Improving Multi Hop Question Answering on Knowledge Graphs}

\author{
    Travis Thompson \\
    North Carolina State University \\
    \texttt{tkthomps@ncsu.edu} \\
    \And
    Seung-Hwan Lim \\
    Oak Ridge National Laboratory \\
    \texttt{lims1@ornl.gov}
    \And
    Paul Liu \\
    North Carolina State University \\
    \texttt{jpliu@ncsu.edu}
    \And
    Ruoying He \\
    North Carolina State University \\
    \texttt{rhe@ncsu.edu}
    \And
    Dongkuan (DK) Xu \\
    North Carolina State University \\
    \texttt{dxu27@ncsu.edu}
}

\begin{document}

\maketitle

\begin{abstract}
Large Language Models (LLMs) have achieved impressive capabilities in language understanding and generation, yet they continue to underperform on knowledge-intensive reasoning tasks due to limited access to structured context and multi-hop information. Retrieval-Augmented Generation (RAG) partially mitigates this by grounding generation in retrieved context, but conventional RAG and GraphRAG methods often fail to capture relational structure across nodes in knowledge graphs. We introduce \textit{Inference-Scaled GraphRAG}, a novel framework that enhances LLM-based graph reasoning by applying inference-time compute scaling. Our method combines sequential scaling with deep chain-of-thought graph traversal, and parallel scaling with majority voting over sampled trajectories within an interleaved reasoning-execution loop. Experiments on the GRBench benchmark demonstrate that our approach significantly improves multi-hop question answering performance, achieving substantial gains over both traditional GraphRAG and prior graph traversal baselines. These findings suggest that inference-time scaling is a practical and architecture-agnostic solution for structured knowledge reasoning with LLMs.
\end{abstract}

\section{Introduction}

Large Language Models (LLMs)~\citep{grattafiori2024llama3herdmodels,jiang2024mixtralexperts,yang2025qwen3technicalreport} have shown remarkable capabilities in natural language understanding and generation, with performance gains primarily driven by scaling self-supervised pretraining~\citep{zhao2025surveylargelanguagemodels}. However, despite these advances, LLMs continue to face challenges on \textit{knowledge-intensive reasoning tasks}~\citep{zheng2024clrfactevaluatingcomplexlogical}, such as answering scientific or domain-specific questions~\citep{tonmoy2024comprehensivesurveyhallucinationmitigation}. These tasks require not only general reasoning abilities but also access to external sources of structured and unstructured information.

To address this, \textit{Retrieval-Augmented Generation (RAG)}~\citep{lewis2020retrieval,gao2024retrievalaugmentedgenerationlargelanguage} has emerged as a widely adopted framework that augments LLMs with access to external text corpora. By retrieving relevant passages and incorporating them as context, RAG improves factual grounding~\citep{10.5555/3495724.3496517} and the quality of generated responses. Nonetheless, RAG suffers from a significant limitation: relevant information is often distributed across multiple text units, and flat retrieval methods struggle to identify and combine information effectively.

\textit{GraphRAG} \citep{edge2025localglobalgraphrag,peng2024graphretrievalaugmentedgenerationsurvey,zhang2025surveygraphretrievalaugmentedgeneration} is a recent extension of RAG that introduces a knowledge graph as the retrieval backbone. Knowledge graphs provide a structured representation of entities and their relationships, enabling retrieval mechanisms to exploit relational dependencies and improve coherence in the retrieved context. By integrating graph-structured data, GraphRAG enhances recall of relevant information and enables more interpretable knowledge augmentation. 

Current GraphRAG implementations \citep{jin2024graph, gao2024retrievalaugmentedgenerationlargelanguage, wei2025instructrag} often neglect \textit{structure context}, treating graph nodes as individual context units ignoring the structural relationships between nodes and their local subgraph. This severely limits performance on tasks requiring synthesis of information across nodes.    
Effective utilization of knowledge graphs in RAG requires not only retrieval of relevant nodes but \textit{explicit traversal} of the graph structure to reason across multiple hops. This is non-trivial for LLMs due to the \textit{graph size explosion problem}, where the number of candidate nodes grows exponentially with traversal depth, overwhelming the context window \citep{liu2023lostmiddlelanguagemodels} and degrading performance.


To address this, we propose a method that applies \textit{inference scaling} \citep{wu2025inference, yao2023tree, yue2025inferencescalinglongcontextretrieval, snell2025scaling} to the problem of graph traversal. Inference scaling, popularized by recent models such as OpenAI o1 \citep{openai2024o1} and DeepSeek R1 \citep{deepseekai2025deepseekr1incentivizingreasoningcapability}, refers to allocating more computational effort at inference time to improve response quality. Our approach leverages both \textit{sequential scaling} \citep{muennighoff2025s1simpletesttimescaling}, where LLMs reason step-by-step in a chain-of-thought \citep{10.5555/3600270.3602070} manner, and \textit{parallel scaling} \citep{brown2024largelanguagemonkeysscaling,wang2023selfconsistency}, where multiple candidate traversals are generated and aggregated via majority voting or best-of-$N$ selection \citep{jinnai2025regularizedbestofnsamplingminimum}.

Concretely, we enable LLMs to perform structured reasoning over knowledge graphs via \textit{interleaved thought-action execution} based on the structure of \textit{Graph-CoT}\citep{jin2024graph}. Traversal is decomposed into three stages: (1) \textit{Reasoning}, where the LLM plans the next step based on the current and previous thoughts; (2) \textit{Interaction}, where the LLM formulates graph queries (e.g., neighbor lookups); and (3) \textit{Execution}, where these queries are performed and results are fed back into the model. This loop enables the model to progressively gather and synthesize multi-hop evidence from the graph.

We empirically evaluate our method on the GRBench \citep{jin2024graph} dataset and observe consistent improvements across multiple model architectures and scales. Our approach surpasses traditional GraphRAG by {\bf 64.7\%}, and exceeds prior graph traversal methods by \textbf{30.3\%}. Furthermore, our method is able to correctly answer \textbf{31.44\%} of difficult multi-hop questions, compared to just \textbf{15.26\%} for baselines.

These results demonstrate that \textit{Inference-Scaled GraphRAG} not only enhances retrieval and reasoning over structured knowledge but also offers a general and architecture-agnostic framework for improving LLM performance on knowledge-intensive tasks.

\begin{figure}[t]
    \centering
    \begin{subfigure}[t]{0.48\textwidth}
        \centering
        \includegraphics[width=\textwidth]{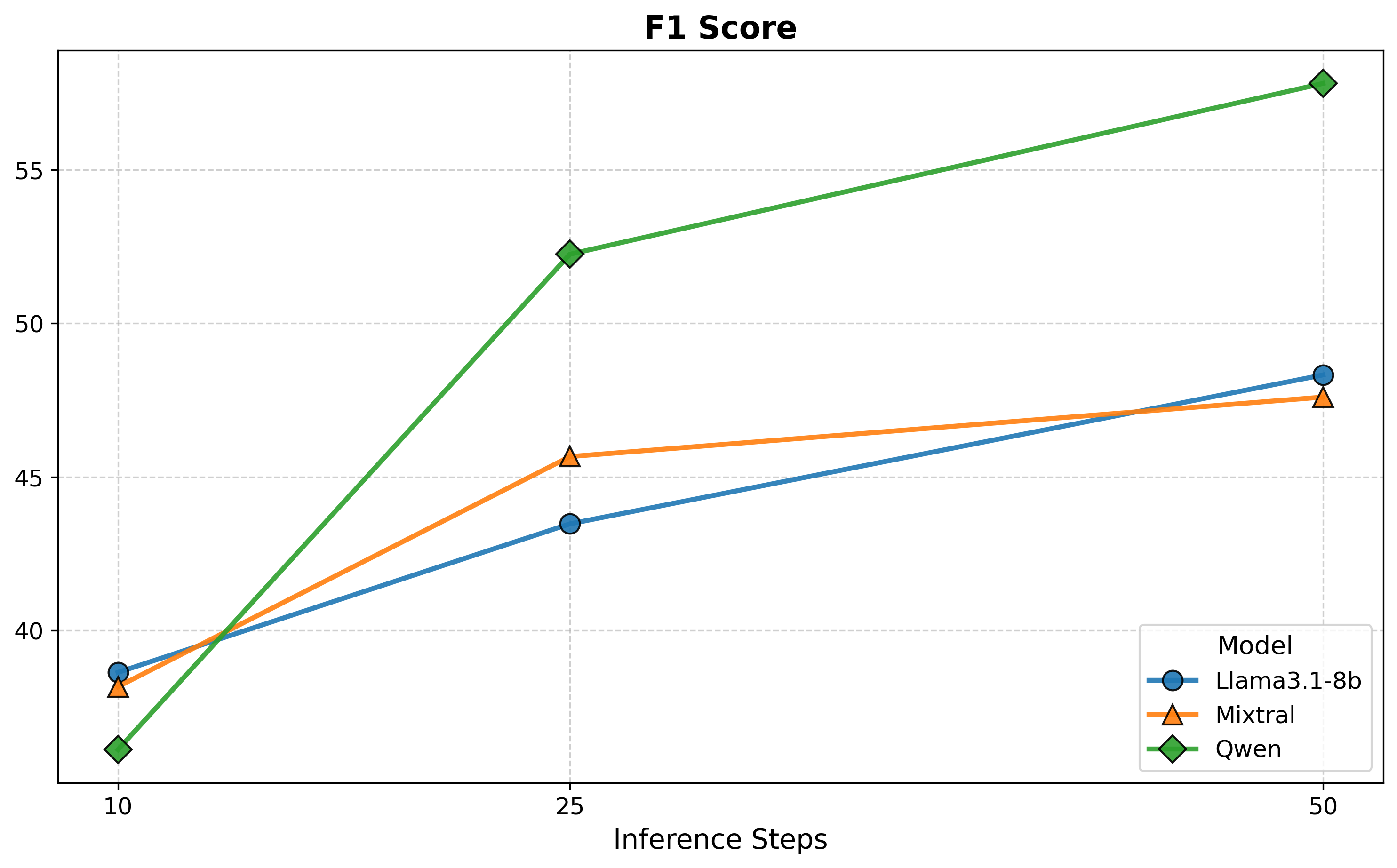}
        \caption{F1 Scores}
        \label{fig:f1_scores}
    \end{subfigure}
    \hfill
    \begin{subfigure}[t]{0.48\textwidth}
        \centering
        \includegraphics[width=\textwidth]{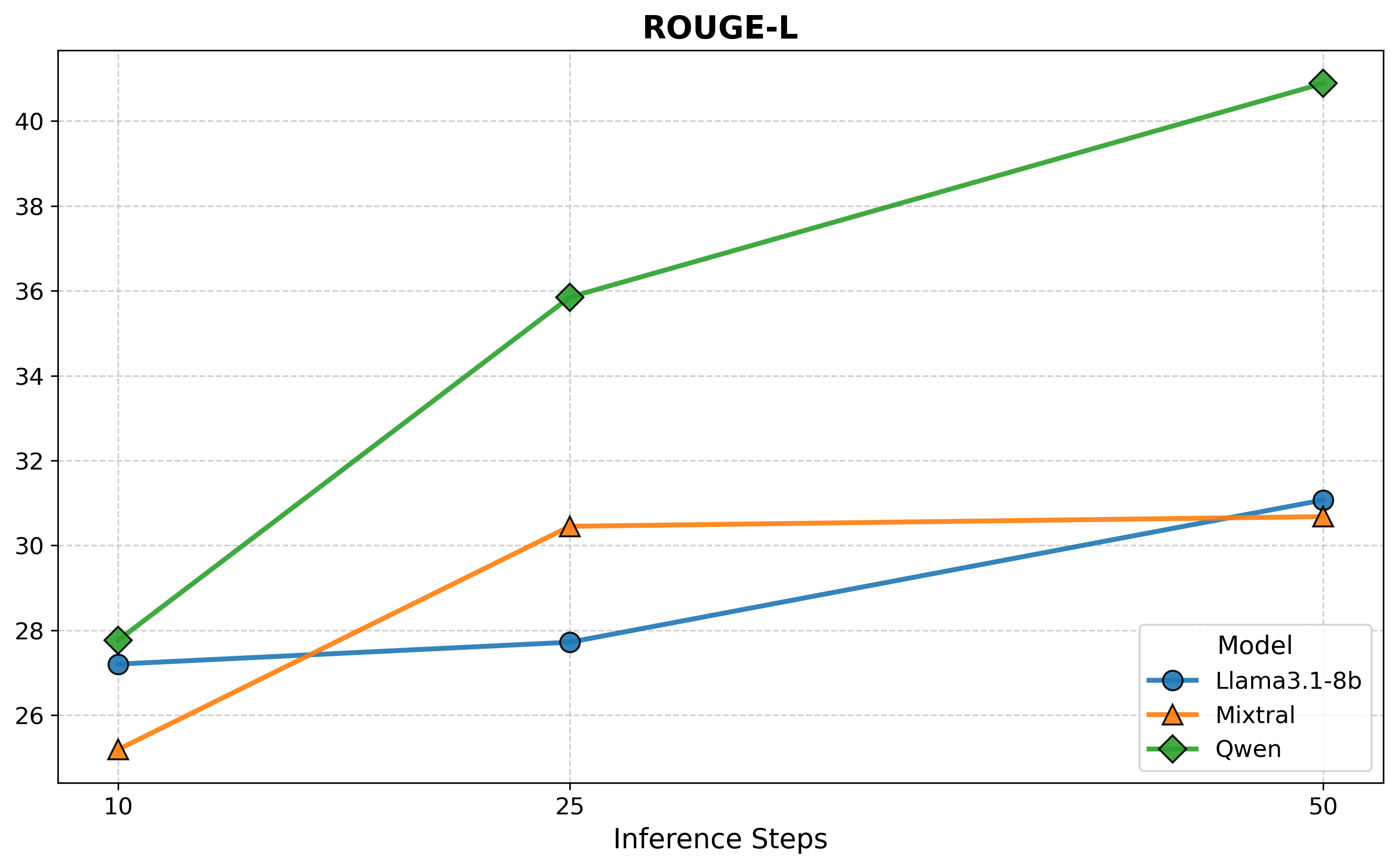}
        \caption{RougeL Scores}
        \label{fig:rougeL_scores}
    \end{subfigure}
    \caption{Performance of \textit{Inference Scaled GraphRAG} under varying inference step counts and model backbones including Llama3.1-8B-Instruct, Mixtral-8x7B-Instruct, and Qwen3-32b. 
    \textbf{(a)} F1 score increases consistently with more inference steps across all model configurations.  
    \textbf{(b)} RougeL scores show a similar trend, reflecting improvements in both symbolic and semantic alignment with ground-truth answers.}
    \label{fig:combined_scores}
\end{figure}
\section{Preliminaries}
\label{preliminaries}

\subsection{Knowledge graphs}

We denote a knowledge graph as \( G = (V, E) \), where \( V \) is the set of nodes (entities) and \( E \) the set of edges (relations). Each node \( v_i \in V \) is associated with feature information \( X_{v_i} \), that may include textual content or metadata. For instance, in an academic domain-specific graph, nodes might represent research papers, with features including authorship, publication venue, or abstract text, while edges capture relationships such as co-authorship or citation links.

\subsection{Retrieval-augmented generation (RAG)}

Large Language Models (LLMs) demonstrate strong generative capabilities but often struggle with reasoning over factual knowledge, especially in specialized domains. Retrieval Augmented Generation (RAG) \citep{lewis2020retrieval} addresses this limitation by incorporating external information retrieval into the generation pipeline. Rather than relying solely on parametric memory, RAG retrieves relevant documents from an external knowledge source and conditions the model’s generation on this context.

RAG implementations typically use dense vector similarity \citep{fan2024surveyragmeetingllms, devlin-etal-2019-bert} to retrieve semantically related documents. While this is effective for recalling isolated facts, it struggles with structured or multi-hop reasoning, particularly when relevant information is distributed across multiple sources \citep{tang2024multihoprag}. GraphRAG \citep{edge2025localglobalgraphrag} extends RAG by using a knowledge graph as the retrieval backbone, allowing the model to leverage the explicit relational structure among entities to guide retrieval and enhance coherence in reasoning.

\subsection{Inference-time scaling}

Traditionally, model performance improvements have been achieved by scaling pretraining \citep{10.5555/3495724.3495883, hoffmann2022trainingcomputeoptimallargelanguage}. Recent studies have shown that allocating additional compute at test time \citep{uesato2022solvingmathwordproblems, 10.5555/3600270.3600548} can significantly improve performance on complex reasoning tasks. Inference-time scaling provides an orthogonal axis for performance improvement without modifying the model architecture.

Two primary paradigms of inference scaling are employed:

\begin{itemize}
    \item \textbf{Sequential Scaling:} The model performs step by step reasoning, with each inference step conditioned on the outputs from previous steps \citep{muennighoff2025s1simpletesttimescaling}. This enables the accumulation of intermediate insights, supporting deeper reasoning and iterative refinement.   

    \item \textbf{Parallel Scaling:} Multiple responses are generated independently \citep{brown2024largelanguagemonkeysscaling} in parallel, and a final output is selected using aggregation strategies such as majority voting or best-of-N. This method improves robustness and mitigates single-sample errors.
\end{itemize}

Our proposed method incorporates both sequential and parallel scaling into the GraphRAG framework, enabling large language models to perform efficient and effective multi-hop reasoning over structured knowledge graphs.

\section{Method}
\label{sec:method}

Retrieval-augmented generation (RAG) \citep{gao2024retrievalaugmentedgenerationlargelanguage, lewis2020retrieval} is a widely used method to integrate large language models (LLMs) with external knowledge sources. In this framework, source documents are embedded and indexed during preprocessing. At inference time, a query is embedded using the same model, and its nearest neighbors are retrieved to form a contextual prompt for the LLM. However, a key limitation of traditional RAG is that relevant information is often distributed across multiple documents. This fragmentation can hinder the LLM’s ability to synthesize a complete and accurate response.

To address this issue, we build on GraphRAG, which leverages a knowledge graph (KG) as the backbone of the retrieval process. In this setting, external knowledge is structured as a graph, and subgraphs, rather than isolated documents, are used as context for the LLM. This representation captures relational structure and enables richer contextualization. Information in KGs often resides in the relationships between nodes, necessitating multi-hop reasoning and graph traversal to answer complex queries. Prior work has shown that LLMs can effectively engage with KGs through iterative reasoning and interaction \citep{jin2024graph}. We extend this paradigm and enhance it with inference-time scaling strategies to improve performance on reasoning-intensive tasks.

\subsection{Graph traversal with LLMs}

Our framework, shown in figure \ref{fig:ISGR_overview}, follows an iterative reasoning and interaction loop between the LLM and the KG, enabling compositional and multi-hop reasoning. Each iteration consists of two phases:


\begin{figure}[t]
\centering
    \includegraphics[width=\textwidth]{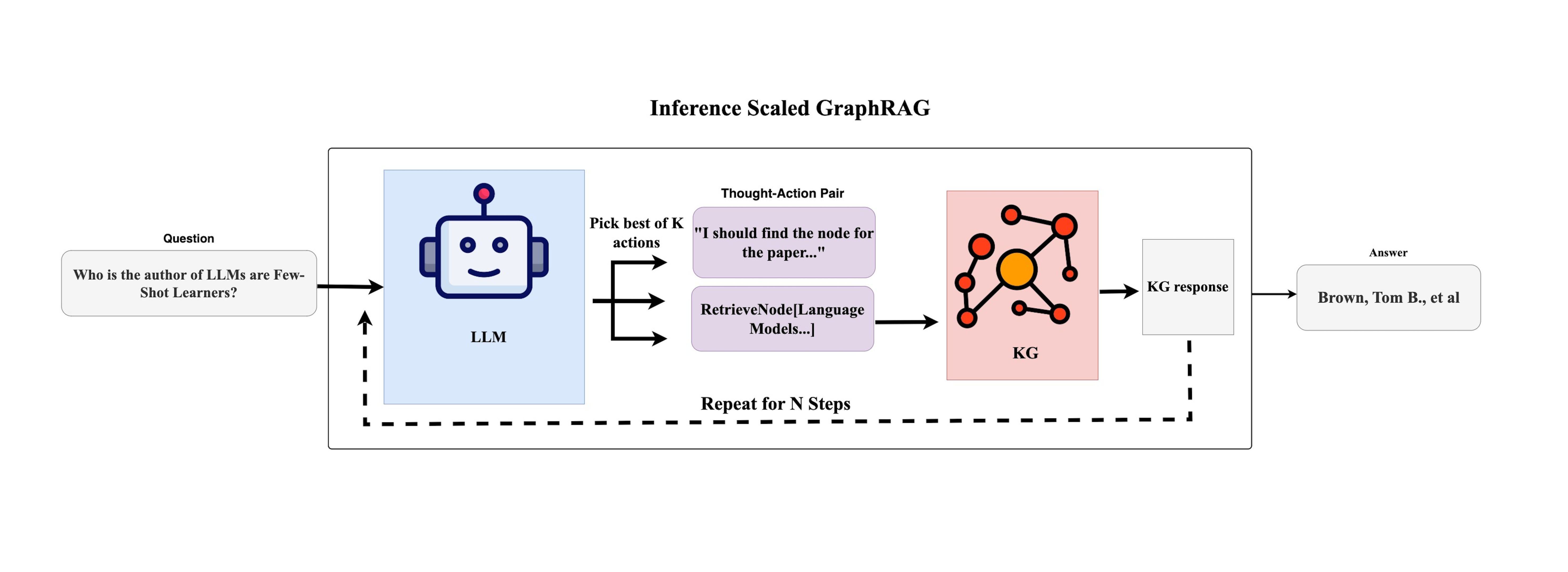}
    \caption{Overview of \textit{Inference Scaled GraphRAG}. The diagram illustrates the iterative process where a question is processed by an LLM. The LLM generates k thought-actions pairs. We perform majority voting to select the best action to interact with the knowledge graph. The response returned from the knowledge graph is fed back as context to the LLM for generation of the next thought-action pair. Generation continues until we reach an answer or hit the maximum step count}
    \label{fig:ISGR_overview}
\end{figure}


\paragraph{Reasoning.} Given a user query and the current context, the LLM first determines whether the question can be answered or whether additional information is required. If more information is needed, the model generates an intermediate goal. For example, for the question “Who are the authors of Language Models are Few-Shot Learners?”, the model may reason: “We need to first locate the node corresponding to the paper titled Language Models are Few-Shot Learners.”

\paragraph{Interaction.} Based on its reasoning, the LLM generates a function call to interact with the KG. We define a fixed set of functions the model can call \citep{yao2023reactsynergizingreasoningacting,jin2024graph}:
\begin{itemize}
\item RetrieveNode: Performs semantic search over the graph to retrieve the node most similar to a given query (equivalent to standard RAG).
\item NodeFeature: Returns the feature associated with a specified node.
\item NeighborCheck: Retrieves the neighbors of a given node along with edge information.
\item NodeDegree: Returns the degree of a node.
\end{itemize}
These functions define an interface between the LLM and the graph environment, effectively framing the LLM as a reasoning agent \citep{xi2023risepotentiallargelanguage, zhuang2023toolqa} operating over a structured space. This agent-like formulation allows us to explicitly control and scale the inference process, as described next.

\subsection{Inference-time scaling}

Multi-hop reasoning is essential in KG-based QA because critical information may be distributed across distant nodes. However, as the number of hops increases, the size of the relevant subgraph grows exponentially. LLMs often struggle with such compositional tasks due to limited context length and inference-time compute. QA performance degrades significantly on questions requiring multi-hop or inductive reasoning.

To address this, we introduce training free \citep{dong2024surveyincontextlearning} inference-time scaling strategies aimed at improving the reasoning ability of LLMs. These strategies are inspired by recent work showing that increasing inference-time compute can substantially improve LLM performance on tasks such as mathematical reasoning and program synthesis \citep{wu2025inference}. We consider two orthogonal axes of inference scaling:

\paragraph{Sequential scaling.} This approach increases the number of reasoning steps the model is allowed to take, thereby enabling deeper graph traversal. By extending the chain-of-thought, the model can accumulate intermediate knowledge, revise earlier assumptions, and explore the graph more effectively. In our setting, we increase the number of reasoning-interaction loops our model can perform before returning a response. This results in the model generating a much longer chain-of-thought and constructing more informed, contextually grounded answers by progressively integrating information across multiple hops in the knowledge graph.

\paragraph{Parallel scaling.} Parallel scaling enhances robustness by sampling multiple independent reasoning trajectories simultaneously and aggregating their outputs. We use majority voting over the sampled thought-action pairs \citep{brown2024largelanguagemonkeysscaling} to select the most consistent action. This approach improves reliability without the need for a learned verifier or extra supervision. In particular, we apply majority voting at the level of \texttt{Interaction} function calls, enabling us to identify and prioritize common reasoning paths across the sampled runs.


\paragraph{Budget forcing.} Our formulation naturally supports budget forcing \citep{muennighoff2025s1simpletesttimescaling}, wherein we impose explicit limits on the number of reasoning steps (sequential scaling) and samples (parallel scaling) allowed during inference. This budget controls the computational cost and aligns with recent trends in controllable inference for LLMs. By tuning these parameters, we can trade off between performance and efficiency in a predictable manner.

\section{Experiments}
We conduct a comprehensive set of experiments to evaluate the effectiveness of our proposed method, \textit{Inference Scaled GraphRAG}, in enhancing multi-hop question answering on knowledge graphs. Specifically, we establish the following:

\begin{itemize}
    \item Inference-time scaling improves performance predictably on graph-based QA. 
    \item Our method improves performance across different levels of reasoning difficulty. We notably see improvements on questions that require multiple hops on the graph. 
    \item Our method performs favorably against traditional GraphRAG as well as previous graph traversal approaches.
\end{itemize}

\subsection{Benchmark dataset}

We evaluate our method on \textbf{GRBench} \citep{jin2024graph}, a knowledge graph-based question answering benchmark designed to assess multi-hop reasoning over structured data. GRBench includes a collection of domain-specific knowledge graphs spanning disciplines such as academic citations, e-commerce, literature, and healthcare. Each knowledge graph is paired with a set of natural language questions that require the model to retrieve and reason over relevant nodes and edges in the graph.

Questions are categorized into three difficulty levels based on the complexity of graph traversal and reasoning required to answer it:

\begin{itemize}
    \item \textbf{Easy}: Can be answered with a single node query or a one-hop traversal.
    \item \textbf{Medium}: Require multi-hop reasoning across several connected entities in the graph.
    \item \textbf{Hard}: Involve inductive or abstract reasoning, often requiring synthesizing multiple paths or patterns across the graph.
\end{itemize}

This structure makes GRBench an ideal benchmark for evaluating inference-time reasoning techniques, as it explicitly tests the model's ability to handle both shallow and deep reasoning tasks across diverse graph structures.

\subsection{Baselines and metrics}

We compare our approach with two categories of baselines:

\paragraph{GraphRAG}
We utilize an embedding model to retrieve semantically similar nodes and their surrounding subgraph. This text is fed to the LLM as context \citep{ye2024language}. We use standard 1 hop retrieval.

\paragraph{GraphCoT.}
the current state-of-the-art method for LLM-based graph traversal using function calls and reasoning steps. We retain their prompting style and retrieval mechanisms, and only vary the underlying model.

We adopt both symbolic and semantic evaluation metrics:

\begin{itemize}
    \item \textbf{RougeL:} Measures overlap based on the longest common subsequence between the generated answer and ground truth.
    \item \textbf{BERTScore (F1):} Uses BERT embeddings to compute token-level similarity between answers and references. We use DistilBERT-base-uncased for embedding generation.
\end{itemize}

\subsection{Implementation details}

All experiments are conducted on NVIDIA a100 and h100 GPUs using Python 3.9 and Huggingface Transformers v4.51.3. We use MPNet-v2 as the retriever and FAISS \citep{johnson2019billion} for indexing. Unless otherwise specified, Llama3.1-8B-Instruct serves as our main language model. We set temperature to 0.7 and utilize top-p sampling with a threshold of 0.9 to balance diversity and coherence in generated responses. Prompting templates and hyperparameters are provided in Appendix A.

We vary two inference scaling parameters to investigate the performance of both sequential and parallel inference scaling:
\begin{itemize}
    \item Number of reasoning steps. Scaling the number of reasoning steps corresponds to scaling inference sequentially. The LLM conditions the current generation on all previous thought action pairs.  
    \item Number of sampled thought-action pairs for majority voting. We utilize majority voting for our parallel scaling implementation in order to isolate the performance of inference scaling rather than utilizing an external reward model for verification.
\end{itemize}






\begin{table*}[t]
\centering
\caption{F1 performance on the different domains of GRBench, comparing: (i) standard GraphRAG; (ii) GraphCoT with no inference scaling (equivalent to the Inference Scaled GraphRAG configuration of 10 steps and 1 vote); and (iii) Inference Scaled GraphRAG with varying step counts and majority voting thresholds. We adopt Llama3.1-8b-instruct as the model backbone for all presented results.}
\label{tab:llama8b-F1-scores}
    \resizebox{\textwidth}{!}{%
        \begin{tabular}{lccccccc}
        \toprule
        Inference config & \textbf{Academic} & \textbf{Amazon} & \textbf{DBLP} & \textbf{Biomedical} & \textbf{goodreads} & \textbf{Legal} & \textbf{Avg} \\
        \midrule
        GraphRAG & 33.71 & 38.34 & 28.01 & 11.29 & 37.26 & 28.56 & 28.87 \\
        \midrule
        10 steps, votes=1  & 41.85 & 47.15 & 34.73 & 8.66 & 48.81 & 32.27 & 36.49 \\
        25 steps, votes=1  & 48.52 & 52.42 & 42.64 & 11.55 & 53.24 & 32.27 & 40.11 \\
        50 steps, votes=1  & 56.27 & 59.62 & 41.71 & 12.26 & 52.94 & 35.33 & 43.02 \\
        \midrule
        10 steps, votes=4  & 47.77 & 41.65 & 47.83 & 12.33 & 53.28 & 34.09 & 39.49 \\
        25 steps, votes=4  & 57.41 & 51.27 & 49.30 & 19.02 & 60.25 & 33.97 & 45.20 \\
        50 steps, votes=4  & 57.76 & 55.24 & 46.00 & 15.58 & 57.35 & 34.46 & 44.67 \\
        \midrule
        10 steps, votes=8  & 47.80 & 49.56 & 42.16 & 14.70 & 54.42 & 36.61 & 40.88 \\
        25 steps, votes=8  & 56.50 & 58.73 & 50.40 & 19.19 & 58.76 & 26.32 & 44.98 \\
        50 steps, votes=8  & 57.79 & \textbf{61.39} & 48.14 & 18.94 & \textbf{61.91} & 34.56 & 47.12 \\
        \midrule
        10 steps, votes=16 & 47.46 & 50.38 & 41.59 & 13.11 & 53.33 & 30.35 & 39.37 \\
        25 steps, votes=16 & 55.34 & 54.26 & \textbf{50.57} & 16.28 & 58.65 & 36.84 & 45.32 \\
        50 steps, votes=16 & \textbf{59.36} & 58.79 & 46.54 & \textbf{19.55} & 61.80 & \textbf{40.88} & \textbf{47.55} \\
        \bottomrule
        \end{tabular}%
     }
\end{table*}

\subsection{Overall performance}

We evaluate the overall effectiveness of Inference Scaled GraphRAG across a diverse set of domains in the GRBench benchmark, measuring F1 and RougeL performance under varying inference scaling configurations. Table~\ref{tab:accuracy_votes_bold} (RougeL scores) and Table~\ref{tab:llama8b-F1-scores} (F1 scores) report these respective results, averaged across nine domains that capture a range of scientific, academic, and general-purpose knowledge graphs.

Our findings demonstrate a clear and consistent trend: increasing the inference budget, either through more reasoning steps (sequential scaling) or by employing majority voting over sampled trajectories (parallel scaling), systematically improves performance across all domains.

Inference Scaled GraphRAG significantly outperforms standard GraphRAG approaches. In terms of RougeL scores (Table~\ref{tab:accuracy_votes_bold}), the standard GraphRAG baseline achieves an average of $6.01$. The base configuration of Inference Scaled GraphRAG (10 steps, 1 vote), which is equivalent to GraphCoT with no inference scaling, achieves an average RougeL score of $27.21$. Performance is further enhanced with increased inference scaling, reaching an average RougeL score of up to $34.33$. 
Similarly, for F1 scores (Table~\ref{tab:llama8b-F1-scores}), Inference Scaled GraphRAG demonstrates substantial improvements. Standard GraphRAG achieves an average F1 score of $28.87$. GraphCoT improves on this with $36.49$, while our method reaches $47.55$ under the highest inference budget. observed in our experiments (configuration: 50 steps, 16 votes). These results underscore the benefits of inference-time compute scaling for enhancing graph-based question answering.

\begin{table*}[t]
    \centering
    \caption{RougeL performance on the different domains of GRBench, comparing: (i) standard GraphRAG; (ii) GraphCoT with no inference scaling (equivalent to the Inference Scaled GraphRAG configuration of 10 steps and 1 vote); and (iii) Inference Scaled GraphRAG with varying step counts and majority voting thresholds. We adopt Llama3.1-8b-instruct as the model backbone for all presented results.}
        \resizebox{\textwidth}{!}{%
        \begin{tabular}{lccccccc}
        \toprule
        Inference config & \textbf{Academic} & \textbf{Amazon} & \textbf{DBLP} & \textbf{Biomedical} & \textbf{Goodreads} & \textbf{Legal} & \textbf{Avg} \\
        \midrule
        GraphRAG & 8.99 & 8.54 & 6.94 & 3.30 & 3.01 & 5.28 & 6.01 \\
        \midrule
        10 steps, votes=1  & 28.53 & 36.78 & 25.08 & 8.15  & 34.26 & 25.11 & 27.21 \\
        25 steps, votes=1  & 30.35 & 36.55 & 26.22 & 5.43  & 33.22 & 24.06 & 27.72 \\
        50 steps, votes=1  & 34.58 & 42.54 & 29.84 & 5.85  & 34.28 & 25.30 & 31.07 \\
        \midrule
        10 steps, votes=4  & 32.43 & 30.27 & 29.38 & 7.39  & 37.26 & 26.45 & 30.00 \\
        25 steps, votes=4  & 35.76 & 34.75 & 34.39 & 11.98 & 40.56 & 24.29 & 32.48 \\
        50 steps, votes=4  & 37.14 & 38.74 & 35.08 & 8.18  & 39.66 & 22.51 & 32.99 \\
        \midrule
        10 steps, votes=8  & 33.79 & 33.83 & 31.14 & 7.69  & 39.15 & 24.88 & 30.56 \\
        25 steps, votes=8  & 36.51 & 41.74 & 32.57 & 11.74 & 39.46 & 16.54 & 32.46 \\
        50 steps, votes=8  & 37.46 & \textbf{44.45} & 33.31 & \textbf{13.08} & 40.13 & 22.99 & 33.82 \\
        \midrule
        10 steps, votes=16 & 32.23 & 36.98 & 28.85 & 8.48  & 40.43 & 20.95 & 29.68 \\
        25 steps, votes=16 & 35.98 & 37.98 & 34.63 & 9.07  & 40.59 & 27.24 & 32.97 \\
        50 steps, votes=16 & \textbf{37.87} & 42.27 & \textbf{36.53} & 10.17 & \textbf{40.89} & \textbf{28.48} & \textbf{34.33} \\
        \bottomrule
        \end{tabular}%
        }
\label{tab:accuracy_votes_bold}
\end{table*}

\subsection{Different LLM backbones}
To understand how our method interacts with different language model sizes and architectures, we evaluate our method with two additional LLM backbones: Mixtral-8x7B-Instruct-v0.1 and Qwen3-32b. These models allow us to assess the generality and robustness of our approach. 

Table \ref{tab:average-f1-rouge} demonstrates that both Mixtral and Qwen3 benefit substantially from increased inference scaling. Performance consistently improves as the number of reasoning steps increases, confirming that deeper graph exploration enhances multi-hop reasoning capabilities irrespective of the underlying LLM backbone.

Notably, Qwen3-32b exhibits the most pronounced gains. While baseline performance across Mixtral-8x7B, Llama3.1-8B, and Qwen3-32b is comparable, with each achieving an F1 score near $37$, Qwen3-32b attains an F1 score approaching $64$ at maximum inference scaling. In contrast, Mixtral’s performance peaks at approximately $49.44$ under the same conditions. This disparity suggests that Qwen3 32b’s stronger inherent reasoning ability enables more effective knowledge graph traversal and information synthesis when augmented with inference time scaling. 

\begin{table*}[ht]
\centering
\caption{Average F1 and RougeL scores for Qwen3-32b and Mixtral-8x7b-Instruct across various configurations.}
\label{tab:average-f1-rouge}
\begin{tabular}{lcccc}
\toprule
& \multicolumn{2}{c}{\textbf{F1 Score}} & \multicolumn{2}{c}{\textbf{RougeL Score}} \\
\textbf{\begin{tabular}[c]{@{}l@{}}Configuration\end{tabular}} & Qwen & Mixtral & Qwen & Mixtral \\
\midrule
10 steps, 1 vote & 36.56 & 38.30 & 27.76 & 25.19 \\
25 steps, 1 vote & 53.00 & 45.88 & 35.85 & 30.45 \\
50 steps, 1 vote & 58.60 & 48.01 & 40.89 & 30.68 \\
\midrule
10 steps, 4 votes & 37.36 & 41.41 & 28.00 & 27.23 \\
25 steps, 4 votes & 52.96 & 45.71 & 36.63 & 28.90 \\
50 steps, 4 votes & 56.98 & 47.58 & 37.99 & 30.32 \\
\midrule
10 steps, 8 votes & 37.65 & 41.86 & 28.15 & 28.33 \\
25 steps, 8 votes & 58.39 & 48.56 & 39.23 & 30.05 \\
50 steps, 8 votes & 61.80 & 46.73 & 42.47 & 31.67 \\
\midrule
10 steps, 16 votes & 40.18 & 43.29 & 30.29 & 28.86 \\
25 steps, 16 votes & 57.98 & 46.94 & 40.11 & 30.36 \\
50 steps, 16 votes & \textbf{63.99} & \textbf{49.44} & \textbf{45.60} & \textbf{32.10} \\
\bottomrule
\end{tabular}
\end{table*}

\subsection{Effect of inference scaling by question difficulty}

\begin{figure}[t]
\centering
    \begin{subfigure}[t]{0.48\textwidth}
        \centering
        \includegraphics[width=\textwidth]{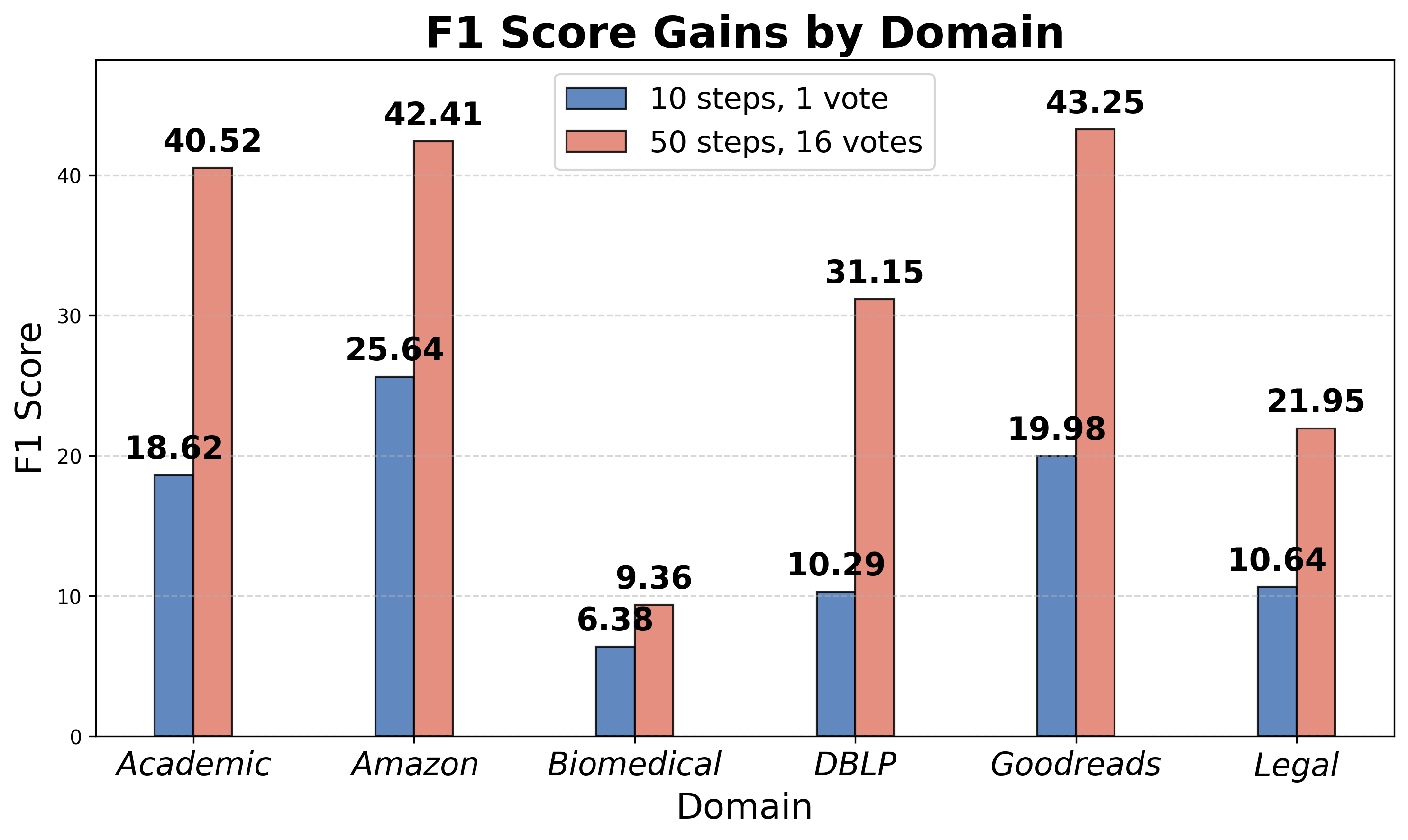}
        \caption{F1 scores on medium and hard questions}
        \label{fig:multi-hopf1}
    \end{subfigure}
    \begin{subfigure}[t]{0.48\textwidth}
        \centering
        \includegraphics[width=\textwidth]{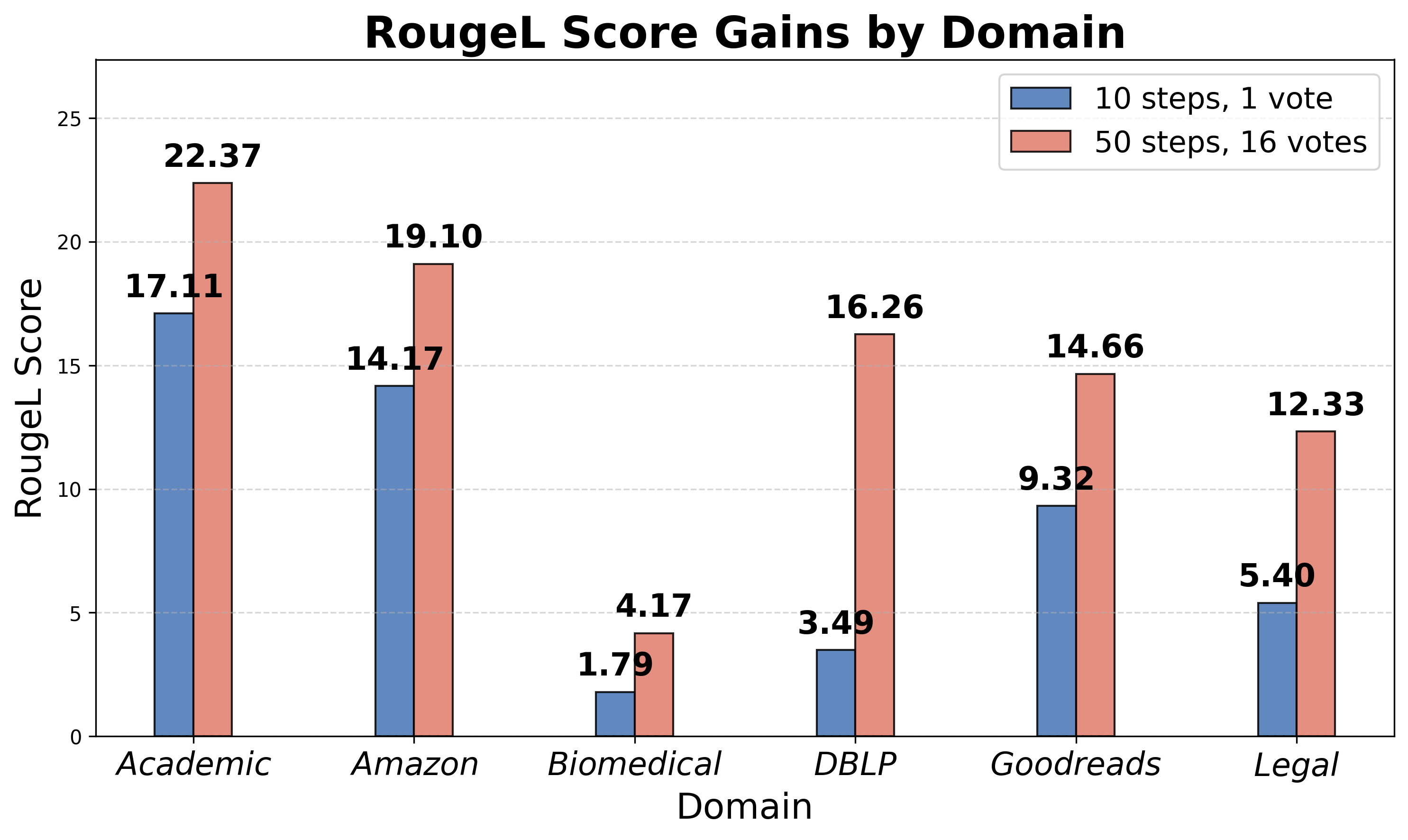}
        \caption{RougeL scores on medium and hard questions}
        \label{fig:multi-hoprougl}
    \end{subfigure}
    \caption{Performance improvements across domains with maximum inference scaling (50 steps, 16 votes) compared to no inference scaling (10 steps, 1 vote). Subfigure (a) reports F1 score gains on medium and hard questions, while (b) shows corresponding improvements in RougeL score. The results demonstrate consistent and substantial gains across all six domains, highlighting the effectiveness of inference scaling for complex question answering }
\end{figure}

To evaluate the impact of inference scaling on multi-hop reasoning, we analyze RougeL and F1 scores on a subset of the questions in GRBench  denoted as {\it medium} and {\it hard}. Figure \ref{fig:multi-hopf1} illustrates the performance improvements achieved by significantly increasing the inference budget (50 steps, 16 votes) over baseline (10 steps, 1 vote). 

We observe that increasing inference compute significantly improves the model's ability to perform multi-hop reasoning. Performance improves by $18.82\%$ on average across domains and $23.27\%$ at most suggesting that with more inference compute the model is able to more effectively explore the graph and retrieve relevant information.

We observe that sequential scaling is the primary driver of performance improvement on these problems, with parallel scaling providing a modest performance boost. We attribute this to the fact that parallel scaling does not directly impact the ability of the model to explore relationships in the graph. For example, a question that can be answered by retrieving relevant information from a node within the local subgraph 3 steps away requires the LLM to generate at least six thought action pairs leaving little room for the LLM to fully explore the subgraph. 

These findings highlight two key takeaways: (1) deeper inference directly improves the multi hop reasoning abilities of LLMs by allowing the model to effectively explore local structure (2) parallel scaling serves as a complementary strategy that improves robustness when applied in conjunction with deep reasoning. Together, these results reinforce the importance of inference time compute allocation in enhancing the symbolic reasoning capacity of LLMs over structured knowledge graphs.

\subsection{Sequential vs parallel scaling}

To better understand the contributions of sequential and parallel inference scaling, we perform an ablation study comparing their isolated and combined effects.

Increasing the number of reasoning steps (i.e. thought action pairs), allows the model to iteratively explore the knowledge graph. This deeper traversal improves multi-hop reasoning by enabling the model to gather and integrate evidence across distant nodes. In table \ref{tab:llama8b-F1-scores}, increasing the step count from 10 to 50 while holding the number of sampled responses constant (votes = 1) led to an average F1 score improvement of $6.53\%$ and a maximum improvement of $14.42\%$, demonstrating the clear benefits of sequential scaling.  

We observe that parallel scaling offers modest performance improvements and primarily boosts robustness against individual errors capturing a wider range of actions. Table \ref{tab:llama8b-F1-scores} shows parallel scaling improves performance by $4.39\%$ on average and $6.86\%$ at most. Parallel scaling is most effective when the base model is already capable of generating high quality reasoning steps. Future work may aim to improve the performance of parallel scaling by utilizing alternative parallel scaling methods such as Best-of-N.


When combined, sequential scaling and parallel scaling yield the highest performance, especially on medium and hard questions requiring multi-hop reasoning. This suggests a synergistic effect where sequential scaling provides the necessary depth exploration and iterative reasoning, while parallel scaling enhances consistency, reliability, and robustness of the final output.

\section{Related Work}

\paragraph{Knowledge Graphs and GraphRAG.}
Knowledge graphs represent entities and their relationships in a structured format, supporting logical reasoning and multi-hop inference. GraphRAG \citep{edge2025localglobalgraphrag} integrates knowledge graphs into the RAG framework by enabling retrieval based on graph structure rather than unstructured text. Recent works have explored using Graph Neural Networks (GNNs) as more sophisticated retrievers \citep{peng2024graphretrievalaugmentedgenerationsurvey, zhang2025surveygraphretrievalaugmentedgeneration}, and enabling LLMs to interact with these GNNs \citep{9046288, wu2025inference}. In addition, works like \citep{he2024gretrieverretrievalaugmentedgenerationtextual} explore methods for incorporating relational structure into retrieval pipelines. Other approaches focus on learning graph node embeddings to enhance retrieval fidelity \citep{devlin-etal-2019-bert}. ToolQA \citep{zhuang2023toolqa} demonstrates the effectiveness of structured tool interfaces, an idea conceptually aligned with function-based graph interaction. Despite these advancements, GraphRAG systems often struggle with the scalability of multi-hop traversal, motivating methods like our own.

\paragraph{Chain-of-Thought and Inference Scaling.}
Inference scaling allows LLMs to perform more effective reasoning by allocating more compute at test time. Researchers have shown that LLMs can learn to allocate more compute through reinforcement learning on verifiable outcomes \citep{lambert2025tulu3pushingfrontiers}. Techniques such as self-consistency \citep{wang2023selfconsistency}, Best-of-N sampling \citep{jinnai2025regularizedbestofnsamplingminimum}, and Tree-of-Thoughts \citep{yao2023tree} improve reasoning robustness and diversity. Recent frameworks like OpenAI o1 \citep{openai2024o1} and DeepSeek R1 \citep{deepseekai2025deepseekr1incentivizingreasoningcapability} demonstrate that test-time scaling can significantly enhance LLM capabilities without altering model weights. The use of budget forcing and controllable compute \citep{muennighoff2025s1simpletesttimescaling} has made it easier to balance cost-performance tradeoffs. Furthermore, Monte Carlo Tree Search strategies \citep{zhang2024rest} offer a path for combining structured planning with sampling-based scaling. Reinforcement learning techniques \citep{gulcehre2023reinforcedselftrainingrestlanguage} show potential for optimizing reasoning strategies in procedural tasks. Scaling laws for inference \citep{wu2025inference} provide theoretical support for compute-efficient design. These inference scaling techniques, while originally designed for symbolic or open-ended tasks, align naturally with the demands of structured graph traversal and motivate our framework’s design.

\paragraph{LLM-based Graph Traversal.}
Recent efforts such as GraphCoT \citep{jin2024graph} explore how LLMs can interact with knowledge graphs using iterative reasoning and function-based graph APIs \citep{jin2024largelanguagemodelsgraphs}. G-Retriever \citep{he2024gretrieverretrievalaugmentedgenerationtextual} focuses on the textual grounding of graph reasoning. Tool-use datasets such as ToolQA \citep{zhuang2023toolqa} and function-agent frameworks \citep{xi2023risepotentiallargelanguage} extend these capabilities to more general interaction settings. In-context learning strategies \citep{dong2024surveyincontextlearning} remain the dominant mechanism for teaching LLMs how to traverse structured environments. However, these methods struggle to generalize to unseen graph topologies. Techniques that enable self-reflection and correction \citep{uesato2022solvingmathwordproblems, 10.5555/3600270.3602070} can aid traversal fidelity. Our work extends this line by explicitly scaling inference compute and structuring the traversal process into reasoning-action-execution loops, allowing for deeper and more accurate reasoning across graph structures.

\section{Conclusion}




We present \textit{Inference Scaled GraphRAG}, a framework for enhancing multi-hop reasoning over knowledge graphs by applying inference-time scaling. By combining \textit{sequential scaling} (multi-step reasoning) and \textit{parallel scaling} (majority voting), our method enables LLMs to more effectively traverse structured contexts.

Empirical results on the GRBench benchmark demonstrate a \textbf{64.7\%} improvement over traditional GraphRAG and a \textbf{30.3\%} improvement over previous graph traversal methods. Furthermore, our method improves performance on difficult multi-hop questions from \textbf{15.26\%} to \textbf{31.44\%} more than doubling performance. 

Our approach validates that inference time scaling is a practical and general solution for structured graph reasoning with LLMs. 

\paragraph{Limitations and future work}
Currently, our majority voting mechanism selects actions based solely on frequency, without evaluating answer correctness. This means incorrect but frequent trajectories may dominate. Future work may explore the use of external verifiers to assess both final answers as well as intermediate results. 

Our framework relies on in-context examples to teach the LLM how to explore the graph. Future work may use reinforcement learning to directly optimize the traversal policy \citep{gulcehre2023reinforcedselftrainingrestlanguage, zhang2024rest}. This would allow the model to receive feedback on successful reasoning chains, improve long-horizon planning, and adapt its exploration strategy based on task rewards leading to more robust and generalizable graph traversal behaviors.

Together, these results position inference-time scaling as a foundational capability for enabling scalable and robust graph reasoning with LLMs. 


\bibliographystyle{plainnat}
\bibliography{references}

\begin{thebibliography}{45}
\providecommand{\natexlab}[1]{#1}
\providecommand{\url}[1]{\texttt{#1}}
\expandafter\ifx\csname urlstyle\endcsname\relax
  \providecommand{\doi}[1]{doi: #1}\else
  \providecommand{\doi}{doi: \begingroup \urlstyle{rm}\Url}\fi

\bibitem[Brown et~al.(2024)Brown, Juravsky, Ehrlich, Clark, Le, Ré, and Mirhoseini]{brown2024largelanguagemonkeysscaling}
Bradley Brown, Jordan Juravsky, Ryan Ehrlich, Ronald Clark, Quoc~V. Le, Christopher Ré, and Azalia Mirhoseini.
\newblock Large language monkeys: Scaling inference compute with repeated sampling, 2024.
\newblock URL \url{https://arxiv.org/abs/2407.21787}.

\bibitem[Brown et~al.(2020)Brown, Mann, Ryder, Subbiah, Kaplan, Dhariwal, Neelakantan, Shyam, Sastry, Askell, Agarwal, Herbert-Voss, Krueger, Henighan, Child, Ramesh, Ziegler, Wu, Winter, Hesse, Chen, Sigler, Litwin, Gray, Chess, Clark, Berner, McCandlish, Radford, Sutskever, and Amodei]{10.5555/3495724.3495883}
Tom~B. Brown, Benjamin Mann, Nick Ryder, Melanie Subbiah, Jared Kaplan, Prafulla Dhariwal, Arvind Neelakantan, Pranav Shyam, Girish Sastry, Amanda Askell, Sandhini Agarwal, Ariel Herbert-Voss, Gretchen Krueger, Tom Henighan, Rewon Child, Aditya Ramesh, Daniel~M. Ziegler, Jeffrey Wu, Clemens Winter, Christopher Hesse, Mark Chen, Eric Sigler, Mateusz Litwin, Scott Gray, Benjamin Chess, Jack Clark, Christopher Berner, Sam McCandlish, Alec Radford, Ilya Sutskever, and Dario Amodei.
\newblock Language models are few-shot learners.
\newblock In \emph{Proceedings of the 34th International Conference on Neural Information Processing Systems}, NIPS '20, Red Hook, NY, USA, 2020. Curran Associates Inc.
\newblock ISBN 9781713829546.

\bibitem[DeepSeek-AI et~al.(2025)DeepSeek-AI, Guo, Yang, Zhang, Song, Zhang, Xu, Zhu, Ma, Wang, Bi, Zhang, Yu, Wu, Wu, Gou, Shao, Li, Gao, Liu, Xue, Wang, Wu, Feng, Lu, Zhao, Deng, Zhang, Ruan, Dai, Chen, Ji, Li, Lin, Dai, Luo, Hao, Chen, Li, Zhang, Bao, Xu, Wang, Ding, Xin, Gao, Qu, Li, Guo, Li, Wang, Chen, Yuan, Qiu, Li, Cai, Ni, Liang, Chen, Dong, Hu, Gao, Guan, Huang, Yu, Wang, Zhang, Zhao, Wang, Zhang, Xu, Xia, Zhang, Zhang, Tang, Li, Wang, Li, Tian, Huang, Zhang, Wang, Chen, Du, Ge, Zhang, Pan, Wang, Chen, Jin, Chen, Lu, Zhou, Chen, Ye, Wang, Yu, Zhou, Pan, Li, Zhou, Wu, Ye, Yun, Pei, Sun, Wang, Zeng, Zhao, Liu, Liang, Gao, Yu, Zhang, Xiao, An, Liu, Wang, Chen, Nie, Cheng, Liu, Xie, Liu, Yang, Li, Su, Lin, Li, Jin, Shen, Chen, Sun, Wang, Song, Zhou, Wang, Shan, Li, Wang, Wei, Zhang, Xu, Li, Zhao, Sun, Wang, Yu, Zhang, Shi, Xiong, He, Piao, Wang, Tan, Ma, Liu, Guo, Ou, Wang, Gong, Zou, He, Xiong, Luo, You, Liu, Zhou, Zhu, Xu, Huang, Li, Zheng, Zhu, Ma, Tang, Zha, Yan, Ren, Ren, Sha, Fu, Xu, Xie, Zhang,
  Hao, Ma, Yan, Wu, Gu, Zhu, Liu, Li, Xie, Song, Pan, Huang, Xu, Zhang, and Zhang]{deepseekai2025deepseekr1incentivizingreasoningcapability}
DeepSeek-AI, Daya Guo, Dejian Yang, Haowei Zhang, Junxiao Song, Ruoyu Zhang, Runxin Xu, Qihao Zhu, Shirong Ma, Peiyi Wang, Xiao Bi, Xiaokang Zhang, Xingkai Yu, Yu~Wu, Z.~F. Wu, Zhibin Gou, Zhihong Shao, Zhuoshu Li, Ziyi Gao, Aixin Liu, Bing Xue, Bingxuan Wang, Bochao Wu, Bei Feng, Chengda Lu, Chenggang Zhao, Chengqi Deng, Chenyu Zhang, Chong Ruan, Damai Dai, Deli Chen, Dongjie Ji, Erhang Li, Fangyun Lin, Fucong Dai, Fuli Luo, Guangbo Hao, Guanting Chen, Guowei Li, H.~Zhang, Han Bao, Hanwei Xu, Haocheng Wang, Honghui Ding, Huajian Xin, Huazuo Gao, Hui Qu, Hui Li, Jianzhong Guo, Jiashi Li, Jiawei Wang, Jingchang Chen, Jingyang Yuan, Junjie Qiu, Junlong Li, J.~L. Cai, Jiaqi Ni, Jian Liang, Jin Chen, Kai Dong, Kai Hu, Kaige Gao, Kang Guan, Kexin Huang, Kuai Yu, Lean Wang, Lecong Zhang, Liang Zhao, Litong Wang, Liyue Zhang, Lei Xu, Leyi Xia, Mingchuan Zhang, Minghua Zhang, Minghui Tang, Meng Li, Miaojun Wang, Mingming Li, Ning Tian, Panpan Huang, Peng Zhang, Qiancheng Wang, Qinyu Chen, Qiushi Du, Ruiqi Ge, Ruisong
  Zhang, Ruizhe Pan, Runji Wang, R.~J. Chen, R.~L. Jin, Ruyi Chen, Shanghao Lu, Shangyan Zhou, Shanhuang Chen, Shengfeng Ye, Shiyu Wang, Shuiping Yu, Shunfeng Zhou, Shuting Pan, S.~S. Li, Shuang Zhou, Shaoqing Wu, Shengfeng Ye, Tao Yun, Tian Pei, Tianyu Sun, T.~Wang, Wangding Zeng, Wanjia Zhao, Wen Liu, Wenfeng Liang, Wenjun Gao, Wenqin Yu, Wentao Zhang, W.~L. Xiao, Wei An, Xiaodong Liu, Xiaohan Wang, Xiaokang Chen, Xiaotao Nie, Xin Cheng, Xin Liu, Xin Xie, Xingchao Liu, Xinyu Yang, Xinyuan Li, Xuecheng Su, Xuheng Lin, X.~Q. Li, Xiangyue Jin, Xiaojin Shen, Xiaosha Chen, Xiaowen Sun, Xiaoxiang Wang, Xinnan Song, Xinyi Zhou, Xianzu Wang, Xinxia Shan, Y.~K. Li, Y.~Q. Wang, Y.~X. Wei, Yang Zhang, Yanhong Xu, Yao Li, Yao Zhao, Yaofeng Sun, Yaohui Wang, Yi~Yu, Yichao Zhang, Yifan Shi, Yiliang Xiong, Ying He, Yishi Piao, Yisong Wang, Yixuan Tan, Yiyang Ma, Yiyuan Liu, Yongqiang Guo, Yuan Ou, Yuduan Wang, Yue Gong, Yuheng Zou, Yujia He, Yunfan Xiong, Yuxiang Luo, Yuxiang You, Yuxuan Liu, Yuyang Zhou, Y.~X. Zhu,
  Yanhong Xu, Yanping Huang, Yaohui Li, Yi~Zheng, Yuchen Zhu, Yunxian Ma, Ying Tang, Yukun Zha, Yuting Yan, Z.~Z. Ren, Zehui Ren, Zhangli Sha, Zhe Fu, Zhean Xu, Zhenda Xie, Zhengyan Zhang, Zhewen Hao, Zhicheng Ma, Zhigang Yan, Zhiyu Wu, Zihui Gu, Zijia Zhu, Zijun Liu, Zilin Li, Ziwei Xie, Ziyang Song, Zizheng Pan, Zhen Huang, Zhipeng Xu, Zhongyu Zhang, and Zhen Zhang.
\newblock Deepseek-r1: Incentivizing reasoning capability in llms via reinforcement learning, 2025.
\newblock URL \url{https://arxiv.org/abs/2501.12948}.

\bibitem[Devlin et~al.(2019)Devlin, Chang, Lee, and Toutanova]{devlin-etal-2019-bert}
Jacob Devlin, Ming-Wei Chang, Kenton Lee, and Kristina Toutanova.
\newblock {BERT}: Pre-training of deep bidirectional transformers for language understanding.
\newblock In Jill Burstein, Christy Doran, and Thamar Solorio, editors, \emph{Proceedings of the 2019 Conference of the North {A}merican Chapter of the Association for Computational Linguistics: Human Language Technologies, Volume 1 (Long and Short Papers)}, pages 4171--4186, Minneapolis, Minnesota, June 2019. Association for Computational Linguistics.
\newblock \doi{10.18653/v1/N19-1423}.
\newblock URL \url{https://aclanthology.org/N19-1423/}.

\bibitem[Dong et~al.(2024)Dong, Li, Dai, Zheng, Ma, Li, Xia, Xu, Wu, Liu, Chang, Sun, Li, and Sui]{dong2024surveyincontextlearning}
Qingxiu Dong, Lei Li, Damai Dai, Ce~Zheng, Jingyuan Ma, Rui Li, Heming Xia, Jingjing Xu, Zhiyong Wu, Tianyu Liu, Baobao Chang, Xu~Sun, Lei Li, and Zhifang Sui.
\newblock A survey on in-context learning, 2024.
\newblock URL \url{https://arxiv.org/abs/2301.00234}.

\bibitem[Edge et~al.(2025)Edge, Trinh, Cheng, Bradley, Chao, Mody, Truitt, Metropolitansky, Ness, and Larson]{edge2025localglobalgraphrag}
Darren Edge, Ha~Trinh, Newman Cheng, Joshua Bradley, Alex Chao, Apurva Mody, Steven Truitt, Dasha Metropolitansky, Robert~Osazuwa Ness, and Jonathan Larson.
\newblock From local to global: A graph rag approach to query-focused summarization, 2025.
\newblock URL \url{https://arxiv.org/abs/2404.16130}.

\bibitem[Fan et~al.(2024)Fan, Ding, Ning, Wang, Li, Yin, Chua, and Li]{fan2024surveyragmeetingllms}
Wenqi Fan, Yujuan Ding, Liangbo Ning, Shijie Wang, Hengyun Li, Dawei Yin, Tat-Seng Chua, and Qing Li.
\newblock A survey on rag meeting llms: Towards retrieval-augmented large language models, 2024.
\newblock URL \url{https://arxiv.org/abs/2405.06211}.

\bibitem[Gao et~al.(2024)Gao, Xiong, Gao, Jia, Pan, Bi, Dai, Sun, Wang, and Wang]{gao2024retrievalaugmentedgenerationlargelanguage}
Yunfan Gao, Yun Xiong, Xinyu Gao, Kangxiang Jia, Jinliu Pan, Yuxi Bi, Yi~Dai, Jiawei Sun, Meng Wang, and Haofen Wang.
\newblock Retrieval-augmented generation for large language models: A survey, 2024.
\newblock URL \url{https://arxiv.org/abs/2312.10997}.

\bibitem[Grattafiori et~al.(2024)Grattafiori, Dubey, Jauhri, Pandey, Kadian, Al-Dahle, Letman, Mathur, Schelten, Vaughan, Yang, Fan, Goyal, Hartshorn, Yang, Mitra, Sravankumar, Korenev, Hinsvark, Rao, Zhang, Rodriguez, Gregerson, Spataru, Roziere, Biron, Tang, Chern, Caucheteux, Nayak, Bi, Marra, McConnell, Keller, Touret, Wu, Wong, Ferrer, Nikolaidis, Allonsius, Song, Pintz, Livshits, Wyatt, Esiobu, Choudhary, Mahajan, Garcia-Olano, Perino, Hupkes, Lakomkin, AlBadawy, Lobanova, Dinan, Smith, Radenovic, Guzmán, Zhang, Synnaeve, Lee, Anderson, Thattai, Nail, Mialon, Pang, Cucurell, Nguyen, Korevaar, Xu, Touvron, Zarov, Ibarra, Kloumann, Misra, Evtimov, Zhang, Copet, Lee, Geffert, Vranes, Park, Mahadeokar, Shah, van~der Linde, Billock, Hong, Lee, Fu, Chi, Huang, Liu, Wang, Yu, Bitton, Spisak, Park, Rocca, Johnstun, Saxe, Jia, Alwala, Prasad, Upasani, Plawiak, Li, Heafield, Stone, El-Arini, Iyer, Malik, Chiu, Bhalla, Lakhotia, Rantala-Yeary, van~der Maaten, Chen, Tan, Jenkins, Martin, Madaan, Malo, Blecher,
  Landzaat, de~Oliveira, Muzzi, Pasupuleti, Singh, Paluri, Kardas, Tsimpoukelli, Oldham, Rita, Pavlova, Kambadur, Lewis, Si, Singh, Hassan, Goyal, Torabi, Bashlykov, Bogoychev, Chatterji, Zhang, Duchenne, Çelebi, Alrassy, Zhang, Li, Vasic, Weng, Bhargava, Dubal, Krishnan, Koura, Xu, He, Dong, Srinivasan, Ganapathy, Calderer, Cabral, Stojnic, Raileanu, Maheswari, Girdhar, Patel, Sauvestre, Polidoro, Sumbaly, Taylor, Silva, Hou, Wang, Hosseini, Chennabasappa, Singh, Bell, Kim, Edunov, Nie, Narang, Raparthy, Shen, Wan, Bhosale, Zhang, Vandenhende, Batra, Whitman, Sootla, Collot, Gururangan, Borodinsky, Herman, Fowler, Sheasha, Georgiou, Scialom, Speckbacher, Mihaylov, Xiao, Karn, Goswami, Gupta, Ramanathan, Kerkez, Gonguet, Do, Vogeti, Albiero, Petrovic, Chu, Xiong, Fu, Meers, Martinet, Wang, Wang, Tan, Xia, Xie, Jia, Wang, Goldschlag, Gaur, Babaei, Wen, Song, Zhang, Li, Mao, Coudert, Yan, Chen, Papakipos, Singh, Srivastava, Jain, Kelsey, Shajnfeld, Gangidi, Victoria, Goldstand, Menon, Sharma, Boesenberg,
  Baevski, Feinstein, Kallet, Sangani, Teo, Yunus, Lupu, Alvarado, Caples, Gu, Ho, Poulton, Ryan, Ramchandani, Dong, Franco, Goyal, Saraf, Chowdhury, Gabriel, Bharambe, Eisenman, Yazdan, James, Maurer, Leonhardi, Huang, Loyd, Paola, Paranjape, Liu, Wu, Ni, Hancock, Wasti, Spence, Stojkovic, Gamido, Montalvo, Parker, Burton, Mejia, Liu, Wang, Kim, Zhou, Hu, Chu, Cai, Tindal, Feichtenhofer, Gao, Civin, Beaty, Kreymer, Li, Adkins, Xu, Testuggine, David, Parikh, Liskovich, Foss, Wang, Le, Holland, Dowling, Jamil, Montgomery, Presani, Hahn, Wood, Le, Brinkman, Arcaute, Dunbar, Smothers, Sun, Kreuk, Tian, Kokkinos, Ozgenel, Caggioni, Kanayet, Seide, Florez, Schwarz, Badeer, Swee, Halpern, Herman, Sizov, Guangyi, Zhang, Lakshminarayanan, Inan, Shojanazeri, Zou, Wang, Zha, Habeeb, Rudolph, Suk, Aspegren, Goldman, Zhan, Damlaj, Molybog, Tufanov, Leontiadis, Veliche, Gat, Weissman, Geboski, Kohli, Lam, Asher, Gaya, Marcus, Tang, Chan, Zhen, Reizenstein, Teboul, Zhong, Jin, Yang, Cummings, Carvill, Shepard, McPhie,
  Torres, Ginsburg, Wang, Wu, U, Saxena, Khandelwal, Zand, Matosich, Veeraraghavan, Michelena, Li, Jagadeesh, Huang, Chawla, Huang, Chen, Garg, A, Silva, Bell, Zhang, Guo, Yu, Moshkovich, Wehrstedt, Khabsa, Avalani, Bhatt, Mankus, Hasson, Lennie, Reso, Groshev, Naumov, Lathi, Keneally, Liu, Seltzer, Valko, Restrepo, Patel, Vyatskov, Samvelyan, Clark, Macey, Wang, Hermoso, Metanat, Rastegari, Bansal, Santhanam, Parks, White, Bawa, Singhal, Egebo, Usunier, Mehta, Laptev, Dong, Cheng, Chernoguz, Hart, Salpekar, Kalinli, Kent, Parekh, Saab, Balaji, Rittner, Bontrager, Roux, Dollar, Zvyagina, Ratanchandani, Yuvraj, Liang, Alao, Rodriguez, Ayub, Murthy, Nayani, Mitra, Parthasarathy, Li, Hogan, Battey, Wang, Howes, Rinott, Mehta, Siby, Bondu, Datta, Chugh, Hunt, Dhillon, Sidorov, Pan, Mahajan, Verma, Yamamoto, Ramaswamy, Lindsay, Lindsay, Feng, Lin, Zha, Patil, Shankar, Zhang, Zhang, Wang, Agarwal, Sajuyigbe, Chintala, Max, Chen, Kehoe, Satterfield, Govindaprasad, Gupta, Deng, Cho, Virk, Subramanian, Choudhury,
  Goldman, Remez, Glaser, Best, Koehler, Robinson, Li, Zhang, Matthews, Chou, Shaked, Vontimitta, Ajayi, Montanez, Mohan, Kumar, Mangla, Ionescu, Poenaru, Mihailescu, Ivanov, Li, Wang, Jiang, Bouaziz, Constable, Tang, Wu, Wang, Wu, Gao, Kleinman, Chen, Hu, Jia, Qi, Li, Zhang, Zhang, Adi, Nam, Yu, Wang, Zhao, Hao, Qian, Li, He, Rait, DeVito, Rosnbrick, Wen, Yang, Zhao, and Ma]{grattafiori2024llama3herdmodels}
Aaron Grattafiori, Abhimanyu Dubey, Abhinav Jauhri, Abhinav Pandey, Abhishek Kadian, Ahmad Al-Dahle, Aiesha Letman, Akhil Mathur, Alan Schelten, Alex Vaughan, Amy Yang, Angela Fan, Anirudh Goyal, Anthony Hartshorn, Aobo Yang, Archi Mitra, Archie Sravankumar, Artem Korenev, Arthur Hinsvark, Arun Rao, Aston Zhang, Aurelien Rodriguez, Austen Gregerson, Ava Spataru, Baptiste Roziere, Bethany Biron, Binh Tang, Bobbie Chern, Charlotte Caucheteux, Chaya Nayak, Chloe Bi, Chris Marra, Chris McConnell, Christian Keller, Christophe Touret, Chunyang Wu, Corinne Wong, Cristian~Canton Ferrer, Cyrus Nikolaidis, Damien Allonsius, Daniel Song, Danielle Pintz, Danny Livshits, Danny Wyatt, David Esiobu, Dhruv Choudhary, Dhruv Mahajan, Diego Garcia-Olano, Diego Perino, Dieuwke Hupkes, Egor Lakomkin, Ehab AlBadawy, Elina Lobanova, Emily Dinan, Eric~Michael Smith, Filip Radenovic, Francisco Guzmán, Frank Zhang, Gabriel Synnaeve, Gabrielle Lee, Georgia~Lewis Anderson, Govind Thattai, Graeme Nail, Gregoire Mialon, Guan Pang,
  Guillem Cucurell, Hailey Nguyen, Hannah Korevaar, Hu~Xu, Hugo Touvron, Iliyan Zarov, Imanol~Arrieta Ibarra, Isabel Kloumann, Ishan Misra, Ivan Evtimov, Jack Zhang, Jade Copet, Jaewon Lee, Jan Geffert, Jana Vranes, Jason Park, Jay Mahadeokar, Jeet Shah, Jelmer van~der Linde, Jennifer Billock, Jenny Hong, Jenya Lee, Jeremy Fu, Jianfeng Chi, Jianyu Huang, Jiawen Liu, Jie Wang, Jiecao Yu, Joanna Bitton, Joe Spisak, Jongsoo Park, Joseph Rocca, Joshua Johnstun, Joshua Saxe, Junteng Jia, Kalyan~Vasuden Alwala, Karthik Prasad, Kartikeya Upasani, Kate Plawiak, Ke~Li, Kenneth Heafield, Kevin Stone, Khalid El-Arini, Krithika Iyer, Kshitiz Malik, Kuenley Chiu, Kunal Bhalla, Kushal Lakhotia, Lauren Rantala-Yeary, Laurens van~der Maaten, Lawrence Chen, Liang Tan, Liz Jenkins, Louis Martin, Lovish Madaan, Lubo Malo, Lukas Blecher, Lukas Landzaat, Luke de~Oliveira, Madeline Muzzi, Mahesh Pasupuleti, Mannat Singh, Manohar Paluri, Marcin Kardas, Maria Tsimpoukelli, Mathew Oldham, Mathieu Rita, Maya Pavlova, Melanie Kambadur,
  Mike Lewis, Min Si, Mitesh~Kumar Singh, Mona Hassan, Naman Goyal, Narjes Torabi, Nikolay Bashlykov, Nikolay Bogoychev, Niladri Chatterji, Ning Zhang, Olivier Duchenne, Onur Çelebi, Patrick Alrassy, Pengchuan Zhang, Pengwei Li, Petar Vasic, Peter Weng, Prajjwal Bhargava, Pratik Dubal, Praveen Krishnan, Punit~Singh Koura, Puxin Xu, Qing He, Qingxiao Dong, Ragavan Srinivasan, Raj Ganapathy, Ramon Calderer, Ricardo~Silveira Cabral, Robert Stojnic, Roberta Raileanu, Rohan Maheswari, Rohit Girdhar, Rohit Patel, Romain Sauvestre, Ronnie Polidoro, Roshan Sumbaly, Ross Taylor, Ruan Silva, Rui Hou, Rui Wang, Saghar Hosseini, Sahana Chennabasappa, Sanjay Singh, Sean Bell, Seohyun~Sonia Kim, Sergey Edunov, Shaoliang Nie, Sharan Narang, Sharath Raparthy, Sheng Shen, Shengye Wan, Shruti Bhosale, Shun Zhang, Simon Vandenhende, Soumya Batra, Spencer Whitman, Sten Sootla, Stephane Collot, Suchin Gururangan, Sydney Borodinsky, Tamar Herman, Tara Fowler, Tarek Sheasha, Thomas Georgiou, Thomas Scialom, Tobias Speckbacher,
  Todor Mihaylov, Tong Xiao, Ujjwal Karn, Vedanuj Goswami, Vibhor Gupta, Vignesh Ramanathan, Viktor Kerkez, Vincent Gonguet, Virginie Do, Vish Vogeti, Vítor Albiero, Vladan Petrovic, Weiwei Chu, Wenhan Xiong, Wenyin Fu, Whitney Meers, Xavier Martinet, Xiaodong Wang, Xiaofang Wang, Xiaoqing~Ellen Tan, Xide Xia, Xinfeng Xie, Xuchao Jia, Xuewei Wang, Yaelle Goldschlag, Yashesh Gaur, Yasmine Babaei, Yi~Wen, Yiwen Song, Yuchen Zhang, Yue Li, Yuning Mao, Zacharie~Delpierre Coudert, Zheng Yan, Zhengxing Chen, Zoe Papakipos, Aaditya Singh, Aayushi Srivastava, Abha Jain, Adam Kelsey, Adam Shajnfeld, Adithya Gangidi, Adolfo Victoria, Ahuva Goldstand, Ajay Menon, Ajay Sharma, Alex Boesenberg, Alexei Baevski, Allie Feinstein, Amanda Kallet, Amit Sangani, Amos Teo, Anam Yunus, Andrei Lupu, Andres Alvarado, Andrew Caples, Andrew Gu, Andrew Ho, Andrew Poulton, Andrew Ryan, Ankit Ramchandani, Annie Dong, Annie Franco, Anuj Goyal, Aparajita Saraf, Arkabandhu Chowdhury, Ashley Gabriel, Ashwin Bharambe, Assaf Eisenman, Azadeh
  Yazdan, Beau James, Ben Maurer, Benjamin Leonhardi, Bernie Huang, Beth Loyd, Beto~De Paola, Bhargavi Paranjape, Bing Liu, Bo~Wu, Boyu Ni, Braden Hancock, Bram Wasti, Brandon Spence, Brani Stojkovic, Brian Gamido, Britt Montalvo, Carl Parker, Carly Burton, Catalina Mejia, Ce~Liu, Changhan Wang, Changkyu Kim, Chao Zhou, Chester Hu, Ching-Hsiang Chu, Chris Cai, Chris Tindal, Christoph Feichtenhofer, Cynthia Gao, Damon Civin, Dana Beaty, Daniel Kreymer, Daniel Li, David Adkins, David Xu, Davide Testuggine, Delia David, Devi Parikh, Diana Liskovich, Didem Foss, Dingkang Wang, Duc Le, Dustin Holland, Edward Dowling, Eissa Jamil, Elaine Montgomery, Eleonora Presani, Emily Hahn, Emily Wood, Eric-Tuan Le, Erik Brinkman, Esteban Arcaute, Evan Dunbar, Evan Smothers, Fei Sun, Felix Kreuk, Feng Tian, Filippos Kokkinos, Firat Ozgenel, Francesco Caggioni, Frank Kanayet, Frank Seide, Gabriela~Medina Florez, Gabriella Schwarz, Gada Badeer, Georgia Swee, Gil Halpern, Grant Herman, Grigory Sizov, Guangyi, Zhang, Guna
  Lakshminarayanan, Hakan Inan, Hamid Shojanazeri, Han Zou, Hannah Wang, Hanwen Zha, Haroun Habeeb, Harrison Rudolph, Helen Suk, Henry Aspegren, Hunter Goldman, Hongyuan Zhan, Ibrahim Damlaj, Igor Molybog, Igor Tufanov, Ilias Leontiadis, Irina-Elena Veliche, Itai Gat, Jake Weissman, James Geboski, James Kohli, Janice Lam, Japhet Asher, Jean-Baptiste Gaya, Jeff Marcus, Jeff Tang, Jennifer Chan, Jenny Zhen, Jeremy Reizenstein, Jeremy Teboul, Jessica Zhong, Jian Jin, Jingyi Yang, Joe Cummings, Jon Carvill, Jon Shepard, Jonathan McPhie, Jonathan Torres, Josh Ginsburg, Junjie Wang, Kai Wu, Kam~Hou U, Karan Saxena, Kartikay Khandelwal, Katayoun Zand, Kathy Matosich, Kaushik Veeraraghavan, Kelly Michelena, Keqian Li, Kiran Jagadeesh, Kun Huang, Kunal Chawla, Kyle Huang, Lailin Chen, Lakshya Garg, Lavender A, Leandro Silva, Lee Bell, Lei Zhang, Liangpeng Guo, Licheng Yu, Liron Moshkovich, Luca Wehrstedt, Madian Khabsa, Manav Avalani, Manish Bhatt, Martynas Mankus, Matan Hasson, Matthew Lennie, Matthias Reso, Maxim
  Groshev, Maxim Naumov, Maya Lathi, Meghan Keneally, Miao Liu, Michael~L. Seltzer, Michal Valko, Michelle Restrepo, Mihir Patel, Mik Vyatskov, Mikayel Samvelyan, Mike Clark, Mike Macey, Mike Wang, Miquel~Jubert Hermoso, Mo~Metanat, Mohammad Rastegari, Munish Bansal, Nandhini Santhanam, Natascha Parks, Natasha White, Navyata Bawa, Nayan Singhal, Nick Egebo, Nicolas Usunier, Nikhil Mehta, Nikolay~Pavlovich Laptev, Ning Dong, Norman Cheng, Oleg Chernoguz, Olivia Hart, Omkar Salpekar, Ozlem Kalinli, Parkin Kent, Parth Parekh, Paul Saab, Pavan Balaji, Pedro Rittner, Philip Bontrager, Pierre Roux, Piotr Dollar, Polina Zvyagina, Prashant Ratanchandani, Pritish Yuvraj, Qian Liang, Rachad Alao, Rachel Rodriguez, Rafi Ayub, Raghotham Murthy, Raghu Nayani, Rahul Mitra, Rangaprabhu Parthasarathy, Raymond Li, Rebekkah Hogan, Robin Battey, Rocky Wang, Russ Howes, Ruty Rinott, Sachin Mehta, Sachin Siby, Sai~Jayesh Bondu, Samyak Datta, Sara Chugh, Sara Hunt, Sargun Dhillon, Sasha Sidorov, Satadru Pan, Saurabh Mahajan,
  Saurabh Verma, Seiji Yamamoto, Sharadh Ramaswamy, Shaun Lindsay, Shaun Lindsay, Sheng Feng, Shenghao Lin, Shengxin~Cindy Zha, Shishir Patil, Shiva Shankar, Shuqiang Zhang, Shuqiang Zhang, Sinong Wang, Sneha Agarwal, Soji Sajuyigbe, Soumith Chintala, Stephanie Max, Stephen Chen, Steve Kehoe, Steve Satterfield, Sudarshan Govindaprasad, Sumit Gupta, Summer Deng, Sungmin Cho, Sunny Virk, Suraj Subramanian, Sy~Choudhury, Sydney Goldman, Tal Remez, Tamar Glaser, Tamara Best, Thilo Koehler, Thomas Robinson, Tianhe Li, Tianjun Zhang, Tim Matthews, Timothy Chou, Tzook Shaked, Varun Vontimitta, Victoria Ajayi, Victoria Montanez, Vijai Mohan, Vinay~Satish Kumar, Vishal Mangla, Vlad Ionescu, Vlad Poenaru, Vlad~Tiberiu Mihailescu, Vladimir Ivanov, Wei Li, Wenchen Wang, Wenwen Jiang, Wes Bouaziz, Will Constable, Xiaocheng Tang, Xiaojian Wu, Xiaolan Wang, Xilun Wu, Xinbo Gao, Yaniv Kleinman, Yanjun Chen, Ye~Hu, Ye~Jia, Ye~Qi, Yenda Li, Yilin Zhang, Ying Zhang, Yossi Adi, Youngjin Nam, Yu, Wang, Yu~Zhao, Yuchen Hao, Yundi
  Qian, Yunlu Li, Yuzi He, Zach Rait, Zachary DeVito, Zef Rosnbrick, Zhaoduo Wen, Zhenyu Yang, Zhiwei Zhao, and Zhiyu Ma.
\newblock The llama 3 herd of models, 2024.
\newblock URL \url{https://arxiv.org/abs/2407.21783}.

\bibitem[Gulcehre et~al.(2023)Gulcehre, Paine, Srinivasan, Konyushkova, Weerts, Sharma, Siddhant, Ahern, Wang, Gu, Macherey, Doucet, Firat, and de~Freitas]{gulcehre2023reinforcedselftrainingrestlanguage}
Caglar Gulcehre, Tom~Le Paine, Srivatsan Srinivasan, Ksenia Konyushkova, Lotte Weerts, Abhishek Sharma, Aditya Siddhant, Alex Ahern, Miaosen Wang, Chenjie Gu, Wolfgang Macherey, Arnaud Doucet, Orhan Firat, and Nando de~Freitas.
\newblock Reinforced self-training (rest) for language modeling, 2023.
\newblock URL \url{https://arxiv.org/abs/2308.08998}.

\bibitem[He et~al.(2024)He, Tian, Sun, Chawla, Laurent, LeCun, Bresson, and Hooi]{he2024gretrieverretrievalaugmentedgenerationtextual}
Xiaoxin He, Yijun Tian, Yifei Sun, Nitesh~V. Chawla, Thomas Laurent, Yann LeCun, Xavier Bresson, and Bryan Hooi.
\newblock G-retriever: Retrieval-augmented generation for textual graph understanding and question answering, 2024.
\newblock URL \url{https://arxiv.org/abs/2402.07630}.

\bibitem[Hoffmann et~al.(2022)Hoffmann, Borgeaud, Mensch, Buchatskaya, Cai, Rutherford, de~Las~Casas, Hendricks, Welbl, Clark, Hennigan, Noland, Millican, van~den Driessche, Damoc, Guy, Osindero, Simonyan, Elsen, Rae, Vinyals, and Sifre]{hoffmann2022trainingcomputeoptimallargelanguage}
Jordan Hoffmann, Sebastian Borgeaud, Arthur Mensch, Elena Buchatskaya, Trevor Cai, Eliza Rutherford, Diego de~Las~Casas, Lisa~Anne Hendricks, Johannes Welbl, Aidan Clark, Tom Hennigan, Eric Noland, Katie Millican, George van~den Driessche, Bogdan Damoc, Aurelia Guy, Simon Osindero, Karen Simonyan, Erich Elsen, Jack~W. Rae, Oriol Vinyals, and Laurent Sifre.
\newblock Training compute-optimal large language models, 2022.
\newblock URL \url{https://arxiv.org/abs/2203.15556}.

\bibitem[Jiang et~al.(2024)Jiang, Sablayrolles, Roux, Mensch, Savary, Bamford, Chaplot, de~las Casas, Hanna, Bressand, Lengyel, Bour, Lample, Lavaud, Saulnier, Lachaux, Stock, Subramanian, Yang, Antoniak, Scao, Gervet, Lavril, Wang, Lacroix, and Sayed]{jiang2024mixtralexperts}
Albert~Q. Jiang, Alexandre Sablayrolles, Antoine Roux, Arthur Mensch, Blanche Savary, Chris Bamford, Devendra~Singh Chaplot, Diego de~las Casas, Emma~Bou Hanna, Florian Bressand, Gianna Lengyel, Guillaume Bour, Guillaume Lample, Lélio~Renard Lavaud, Lucile Saulnier, Marie-Anne Lachaux, Pierre Stock, Sandeep Subramanian, Sophia Yang, Szymon Antoniak, Teven~Le Scao, Théophile Gervet, Thibaut Lavril, Thomas Wang, Timothée Lacroix, and William~El Sayed.
\newblock Mixtral of experts, 2024.
\newblock URL \url{https://arxiv.org/abs/2401.04088}.

\bibitem[Jin et~al.(2024{\natexlab{a}})Jin, Liu, Han, Jiang, Ji, and Han]{jin2024largelanguagemodelsgraphs}
Bowen Jin, Gang Liu, Chi Han, Meng Jiang, Heng Ji, and Jiawei Han.
\newblock Large language models on graphs: A comprehensive survey, 2024{\natexlab{a}}.
\newblock URL \url{https://arxiv.org/abs/2312.02783}.

\bibitem[Jin et~al.(2024{\natexlab{b}})Jin, Xie, Zhang, Roy, Zhang, Li, Li, Tang, Wang, Meng, and Han]{jin2024graph}
Bowen Jin, Chulin Xie, Jiawei Zhang, Kashob~Kumar Roy, Yu~Zhang, Zheng Li, Ruirui Li, Xianfeng Tang, Suhang Wang, Yu~Meng, and Jiawei Han.
\newblock Graph chain-of-thought: Augmenting large language models by reasoning on graphs.
\newblock In \emph{Findings of the Association for Computational Linguistics: ACL 2024}, pages 163--184, Bangkok, Thailand, 2024{\natexlab{b}}. Association for Computational Linguistics.
\newblock \doi{10.18653/v1/2024.findings-acl.11}.
\newblock URL \url{https://aclanthology.org/2024.findings-acl.11/}.

\bibitem[Jinnai et~al.(2025)Jinnai, Morimura, Ariu, and Abe]{jinnai2025regularizedbestofnsamplingminimum}
Yuu Jinnai, Tetsuro Morimura, Kaito Ariu, and Kenshi Abe.
\newblock Regularized best-of-n sampling with minimum bayes risk objective for language model alignment, 2025.
\newblock URL \url{https://arxiv.org/abs/2404.01054}.

\bibitem[Johnson et~al.(2019)Johnson, Douze, and Jégou]{johnson2019billion}
Jeff Johnson, Matthijs Douze, and Hervé Jégou.
\newblock Billion-scale similarity search with gpus.
\newblock \emph{IEEE Transactions on Big Data}, 7\penalty0 (3):\penalty0 535--547, 2019.
\newblock \doi{10.1109/TBDATA.2019.2921572}.

\bibitem[Lambert et~al.(2025)Lambert, Morrison, Pyatkin, Huang, Ivison, Brahman, Miranda, Liu, Dziri, Lyu, Gu, Malik, Graf, Hwang, Yang, Bras, Tafjord, Wilhelm, Soldaini, Smith, Wang, Dasigi, and Hajishirzi]{lambert2025tulu3pushingfrontiers}
Nathan Lambert, Jacob Morrison, Valentina Pyatkin, Shengyi Huang, Hamish Ivison, Faeze Brahman, Lester James~V. Miranda, Alisa Liu, Nouha Dziri, Shane Lyu, Yuling Gu, Saumya Malik, Victoria Graf, Jena~D. Hwang, Jiangjiang Yang, Ronan~Le Bras, Oyvind Tafjord, Chris Wilhelm, Luca Soldaini, Noah~A. Smith, Yizhong Wang, Pradeep Dasigi, and Hannaneh Hajishirzi.
\newblock Tulu 3: Pushing frontiers in open language model post-training, 2025.
\newblock URL \url{https://arxiv.org/abs/2411.15124}.

\bibitem[Lewis et~al.(2020{\natexlab{a}})Lewis, Perez, Piktus, Petroni, Karpukhin, Goyal, K\"{u}ttler, Lewis, Yih, Rockt\"{a}schel, Riedel, and Kiela]{10.5555/3495724.3496517}
Patrick Lewis, Ethan Perez, Aleksandra Piktus, Fabio Petroni, Vladimir Karpukhin, Naman Goyal, Heinrich K\"{u}ttler, Mike Lewis, Wen-tau Yih, Tim Rockt\"{a}schel, Sebastian Riedel, and Douwe Kiela.
\newblock Retrieval-augmented generation for knowledge-intensive nlp tasks.
\newblock In \emph{Proceedings of the 34th International Conference on Neural Information Processing Systems}, NIPS '20, Red Hook, NY, USA, 2020{\natexlab{a}}. Curran Associates Inc.
\newblock ISBN 9781713829546.

\bibitem[Lewis et~al.(2020{\natexlab{b}})Lewis, Perez, Piktus, Petroni, Karpukhin, Goyal, K{\"u}ttler, Lewis, Yih, Rockt{\"a}schel, et~al.]{lewis2020retrieval}
Patrick Lewis, Ethan Perez, Aleksandra Piktus, Fabio Petroni, Vladimir Karpukhin, Naman Goyal, Heinrich K{\"u}ttler, Mike Lewis, Wen-tau Yih, Tim Rockt{\"a}schel, et~al.
\newblock Retrieval-augmented generation for knowledge-intensive nlp tasks.
\newblock In \emph{Advances in Neural Information Processing Systems}, volume~33, pages 9459--9474, 2020{\natexlab{b}}.

\bibitem[Lewkowycz et~al.(2022)Lewkowycz, Andreassen, Dohan, Dyer, Michalewski, Ramasesh, Slone, Anil, Schlag, Gutman-Solo, Wu, Neyshabur, Gur-Ari, and Misra]{10.5555/3600270.3600548}
Aitor Lewkowycz, Anders Andreassen, David Dohan, Ethan Dyer, Henryk Michalewski, Vinay Ramasesh, Ambrose Slone, Cem Anil, Imanol Schlag, Theo Gutman-Solo, Yuhuai Wu, Behnam Neyshabur, Guy Gur-Ari, and Vedant Misra.
\newblock Solving quantitative reasoning problems with language models.
\newblock In \emph{Proceedings of the 36th International Conference on Neural Information Processing Systems}, NIPS '22, Red Hook, NY, USA, 2022. Curran Associates Inc.
\newblock ISBN 9781713871088.

\bibitem[Liu et~al.(2023)Liu, Lin, Hewitt, Paranjape, Bevilacqua, Petroni, and Liang]{liu2023lostmiddlelanguagemodels}
Nelson~F. Liu, Kevin Lin, John Hewitt, Ashwin Paranjape, Michele Bevilacqua, Fabio Petroni, and Percy Liang.
\newblock Lost in the middle: How language models use long contexts, 2023.
\newblock URL \url{https://arxiv.org/abs/2307.03172}.

\bibitem[Muennighoff et~al.(2025)Muennighoff, Yang, Shi, Li, Fei-Fei, Hajishirzi, Zettlemoyer, Liang, Candès, and Hashimoto]{muennighoff2025s1simpletesttimescaling}
Niklas Muennighoff, Zitong Yang, Weijia Shi, Xiang~Lisa Li, Li~Fei-Fei, Hannaneh Hajishirzi, Luke Zettlemoyer, Percy Liang, Emmanuel Candès, and Tatsunori Hashimoto.
\newblock s1: Simple test-time scaling, 2025.
\newblock URL \url{https://arxiv.org/abs/2501.19393}.

\bibitem[OpenAI(2024)]{openai2024o1}
OpenAI.
\newblock Openai o1 system card, 2024.
\newblock URL \url{https://arxiv.org/abs/2412.16720}.

\bibitem[Peng et~al.(2024)Peng, Zhu, Liu, Bo, Shi, Hong, Zhang, and Tang]{peng2024graphretrievalaugmentedgenerationsurvey}
Boci Peng, Yun Zhu, Yongchao Liu, Xiaohe Bo, Haizhou Shi, Chuntao Hong, Yan Zhang, and Siliang Tang.
\newblock Graph retrieval-augmented generation: A survey, 2024.
\newblock URL \url{https://arxiv.org/abs/2408.08921}.

\bibitem[Snell et~al.(2025)Snell, Lee, Xu, and Kumar]{snell2025scaling}
Charlie~Victor Snell, Jaehoon Lee, Kelvin Xu, and Aviral Kumar.
\newblock Scaling {LLM} test-time compute optimally can be more effective than scaling parameters for reasoning.
\newblock In \emph{The Thirteenth International Conference on Learning Representations}, 2025.
\newblock URL \url{https://openreview.net/forum?id=4FWAwZtd2n}.

\bibitem[Tang and Yang(2024)]{tang2024multihoprag}
Yixuan Tang and Yi~Yang.
\newblock Multihop-{RAG}: Benchmarking retrieval-augmented generation for multi-hop queries.
\newblock In \emph{First Conference on Language Modeling}, 2024.
\newblock URL \url{https://openreview.net/forum?id=t4eB3zYWBK}.

\bibitem[Tonmoy et~al.(2024)Tonmoy, Zaman, Jain, Rani, Rawte, Chadha, and Das]{tonmoy2024comprehensivesurveyhallucinationmitigation}
S.~M Towhidul~Islam Tonmoy, S~M~Mehedi Zaman, Vinija Jain, Anku Rani, Vipula Rawte, Aman Chadha, and Amitava Das.
\newblock A comprehensive survey of hallucination mitigation techniques in large language models, 2024.
\newblock URL \url{https://arxiv.org/abs/2401.01313}.

\bibitem[Uesato et~al.(2022)Uesato, Kushman, Kumar, Song, Siegel, Wang, Creswell, Irving, and Higgins]{uesato2022solvingmathwordproblems}
Jonathan Uesato, Nate Kushman, Ramana Kumar, Francis Song, Noah Siegel, Lisa Wang, Antonia Creswell, Geoffrey Irving, and Irina Higgins.
\newblock Solving math word problems with process- and outcome-based feedback, 2022.
\newblock URL \url{https://arxiv.org/abs/2211.14275}.

\bibitem[Wang et~al.(2023)Wang, Wei, Schuurmans, Le, Chi, Narang, Chowdhery, and Zhou]{wang2023selfconsistency}
Xuezhi Wang, Jason Wei, Dale Schuurmans, Quoc~V Le, Ed~H. Chi, Sharan Narang, Aakanksha Chowdhery, and Denny Zhou.
\newblock Self-consistency improves chain of thought reasoning in language models.
\newblock In \emph{The Eleventh International Conference on Learning Representations}, 2023.
\newblock URL \url{https://openreview.net/forum?id=1PL1NIMMrw}.

\bibitem[Wei et~al.(2022)Wei, Wang, Schuurmans, Bosma, Ichter, Xia, Chi, Le, and Zhou]{10.5555/3600270.3602070}
Jason Wei, Xuezhi Wang, Dale Schuurmans, Maarten Bosma, Brian Ichter, Fei Xia, Ed~H. Chi, Quoc~V. Le, and Denny Zhou.
\newblock Chain-of-thought prompting elicits reasoning in large language models.
\newblock In \emph{Proceedings of the 36th International Conference on Neural Information Processing Systems}, NIPS '22, Red Hook, NY, USA, 2022. Curran Associates Inc.
\newblock ISBN 9781713871088.

\bibitem[Wei et~al.(2025)Wei, Chen, and Meng]{wei2025instructrag}
Zhepei Wei, Wei-Lin Chen, and Yu~Meng.
\newblock Instruct{RAG}: Instructing retrieval-augmented generation via self-synthesized rationales.
\newblock In \emph{The Thirteenth International Conference on Learning Representations}, 2025.
\newblock URL \url{https://openreview.net/forum?id=P1qhkp8gQT}.

\bibitem[Wu et~al.(2025)Wu, Sun, Li, Welleck, and Yang]{wu2025inference}
Yangzhen Wu, Zhiqing Sun, Shanda Li, Sean Welleck, and Yiming Yang.
\newblock Inference scaling laws: An empirical analysis of compute-optimal inference for {LLM} problem-solving.
\newblock In \emph{The Thirteenth International Conference on Learning Representations}, 2025.
\newblock URL \url{https://openreview.net/forum?id=VNckp7JEHn}.

\bibitem[Wu et~al.(2021)Wu, Pan, Chen, Long, Zhang, and Yu]{9046288}
Zonghan Wu, Shirui Pan, Fengwen Chen, Guodong Long, Chengqi Zhang, and Philip~S. Yu.
\newblock A comprehensive survey on graph neural networks.
\newblock \emph{IEEE Transactions on Neural Networks and Learning Systems}, 32\penalty0 (1):\penalty0 4--24, 2021.
\newblock \doi{10.1109/TNNLS.2020.2978386}.

\bibitem[Xi et~al.(2023)Xi, Chen, Guo, He, Ding, Hong, Zhang, Wang, Jin, Zhou, Zheng, Fan, Wang, Xiong, Zhou, Wang, Jiang, Zou, Liu, Yin, Dou, Weng, Cheng, Zhang, Qin, Zheng, Qiu, Huang, and Gui]{xi2023risepotentiallargelanguage}
Zhiheng Xi, Wenxiang Chen, Xin Guo, Wei He, Yiwen Ding, Boyang Hong, Ming Zhang, Junzhe Wang, Senjie Jin, Enyu Zhou, Rui Zheng, Xiaoran Fan, Xiao Wang, Limao Xiong, Yuhao Zhou, Weiran Wang, Changhao Jiang, Yicheng Zou, Xiangyang Liu, Zhangyue Yin, Shihan Dou, Rongxiang Weng, Wensen Cheng, Qi~Zhang, Wenjuan Qin, Yongyan Zheng, Xipeng Qiu, Xuanjing Huang, and Tao Gui.
\newblock The rise and potential of large language model based agents: A survey, 2023.
\newblock URL \url{https://arxiv.org/abs/2309.07864}.

\bibitem[Yang et~al.(2025)Yang, Li, Yang, Zhang, Hui, Zheng, Yu, Gao, Huang, Lv, Zheng, Liu, Zhou, Huang, Hu, Ge, Wei, Lin, Tang, Yang, Tu, Zhang, Yang, Yang, Zhou, Zhou, Lin, Dang, Bao, Yang, Yu, Deng, Li, Xue, Li, Zhang, Wang, Zhu, Men, Gao, Liu, Luo, Li, Tang, Yin, Ren, Wang, Zhang, Ren, Fan, Su, Zhang, Zhang, Wan, Liu, Wang, Cui, Zhang, Zhou, and Qiu]{yang2025qwen3technicalreport}
An~Yang, Anfeng Li, Baosong Yang, Beichen Zhang, Binyuan Hui, Bo~Zheng, Bowen Yu, Chang Gao, Chengen Huang, Chenxu Lv, Chujie Zheng, Dayiheng Liu, Fan Zhou, Fei Huang, Feng Hu, Hao Ge, Haoran Wei, Huan Lin, Jialong Tang, Jian Yang, Jianhong Tu, Jianwei Zhang, Jianxin Yang, Jiaxi Yang, Jing Zhou, Jingren Zhou, Junyang Lin, Kai Dang, Keqin Bao, Kexin Yang, Le~Yu, Lianghao Deng, Mei Li, Mingfeng Xue, Mingze Li, Pei Zhang, Peng Wang, Qin Zhu, Rui Men, Ruize Gao, Shixuan Liu, Shuang Luo, Tianhao Li, Tianyi Tang, Wenbiao Yin, Xingzhang Ren, Xinyu Wang, Xinyu Zhang, Xuancheng Ren, Yang Fan, Yang Su, Yichang Zhang, Yinger Zhang, Yu~Wan, Yuqiong Liu, Zekun Wang, Zeyu Cui, Zhenru Zhang, Zhipeng Zhou, and Zihan Qiu.
\newblock Qwen3 technical report, 2025.
\newblock URL \url{https://arxiv.org/abs/2505.09388}.

\bibitem[Yao et~al.(2023{\natexlab{a}})Yao, Astuti, Peng, Chen, Nijkamp, and Levine]{yao2023tree}
Shunyu Yao, Dian Astuti, Bo~Peng, Danfei Chen, Erik Nijkamp, and Sergey Levine.
\newblock {Tree of Thoughts: Deliberate Problem Solving with Large Language Models}.
\newblock In \emph{Thirty-seventh Conference on Neural Information Processing Systems}, 2023{\natexlab{a}}.
\newblock URL \url{https://openreview.net/forum?id=A12pZ0d5tN}.

\bibitem[Yao et~al.(2023{\natexlab{b}})Yao, Zhao, Yu, Du, Shafran, Narasimhan, and Cao]{yao2023reactsynergizingreasoningacting}
Shunyu Yao, Jeffrey Zhao, Dian Yu, Nan Du, Izhak Shafran, Karthik Narasimhan, and Yuan Cao.
\newblock React: Synergizing reasoning and acting in language models, 2023{\natexlab{b}}.
\newblock URL \url{https://arxiv.org/abs/2210.03629}.

\bibitem[Ye et~al.(2024)Ye, Zhang, Wang, Xu, and Zhang]{ye2024language}
Ruosong Ye, Caiqi Zhang, Runhui Wang, Shuyuan Xu, and Yongfeng Zhang.
\newblock Language is all a graph needs.
\newblock In \emph{Findings of the Association for Computational Linguistics: EACL 2024}, pages 1955--1973, St. Julian's, Malta, March 2024. Association for Computational Linguistics.
\newblock \doi{10.18653/v1/2024.findings-eacl.132}.
\newblock URL \url{https://aclanthology.org/2024.findings-eacl.132/}.

\bibitem[Yue et~al.(2025)Yue, Zhuang, Bai, Hui, Jagerman, Zeng, Qin, Wang, Wang, and Bendersky]{yue2025inferencescalinglongcontextretrieval}
Zhenrui Yue, Honglei Zhuang, Aijun Bai, Kai Hui, Rolf Jagerman, Hansi Zeng, Zhen Qin, Dong Wang, Xuanhui Wang, and Michael Bendersky.
\newblock Inference scaling for long-context retrieval augmented generation, 2025.
\newblock URL \url{https://arxiv.org/abs/2410.04343}.

\bibitem[Zhang et~al.(2024)Zhang, Zhoubian, Hu, Yue, Dong, and Tang]{zhang2024rest}
Dan Zhang, Sining Zhoubian, Ziniu Hu, Yisong Yue, Yuxiao Dong, and Jie Tang.
\newblock Rest-mcts*: Llm self-training via process reward guided tree search.
\newblock \emph{arXiv preprint arXiv:2406.03816}, 2024.

\bibitem[Zhang et~al.(2025)Zhang, Chen, Bei, Yuan, Zhou, Hong, Dong, Chen, Chang, and Huang]{zhang2025surveygraphretrievalaugmentedgeneration}
Qinggang Zhang, Shengyuan Chen, Yuanchen Bei, Zheng Yuan, Huachi Zhou, Zijin Hong, Junnan Dong, Hao Chen, Yi~Chang, and Xiao Huang.
\newblock A survey of graph retrieval-augmented generation for customized large language models, 2025.
\newblock URL \url{https://arxiv.org/abs/2501.13958}.

\bibitem[Zhao et~al.(2025)Zhao, Zhou, Li, Tang, Wang, Hou, Min, Zhang, Zhang, Dong, Du, Yang, Chen, Chen, Jiang, Ren, Li, Tang, Liu, Liu, Nie, and Wen]{zhao2025surveylargelanguagemodels}
Wayne~Xin Zhao, Kun Zhou, Junyi Li, Tianyi Tang, Xiaolei Wang, Yupeng Hou, Yingqian Min, Beichen Zhang, Junjie Zhang, Zican Dong, Yifan Du, Chen Yang, Yushuo Chen, Zhipeng Chen, Jinhao Jiang, Ruiyang Ren, Yifan Li, Xinyu Tang, Zikang Liu, Peiyu Liu, Jian-Yun Nie, and Ji-Rong Wen.
\newblock A survey of large language models, 2025.
\newblock URL \url{https://arxiv.org/abs/2303.18223}.

\bibitem[Zheng et~al.(2024)Zheng, Bai, Wang, Fang, Guo, Yim, and Song]{zheng2024clrfactevaluatingcomplexlogical}
Tianshi Zheng, Jiaxin Bai, Yicheng Wang, Tianqing Fang, Yue Guo, Yauwai Yim, and Yangqiu Song.
\newblock Clr-fact: Evaluating the complex logical reasoning capability of large language models over factual knowledge, 2024.
\newblock URL \url{https://arxiv.org/abs/2407.20564}.

\bibitem[Zhuang et~al.(2023)Zhuang, Yu, Wang, Sun, and Zhang]{zhuang2023toolqa}
Yuchen Zhuang, Yue Yu, Kuan Wang, Haotian Sun, and Chao Zhang.
\newblock Toolqa: A dataset for llm question answering with external tools.
\newblock In \emph{Advances in Neural Information Processing Systems}, volume~36, pages 50117--50143, 2023.

\end{thebibliography}

\newpage
\appendix
\section*{Appendix}
\addcontentsline{toc}{section}{Appendix}
\renewcommand{\thesubsection}{A.\arabic{subsection}}
\renewcommand{\thesubsubsection}{A.\arabic{subsection}.\arabic{subsubsection}}

\section{Prompting templates}
Our prompt is composed of a graph definition, function descriptions, and demonstrations. The complete prompt is shown below, where \texttt{\{graph\_definition\}}, \texttt{\{function\_definitions\}}, and \texttt{\{demonstrations\}} correspond to each respective section in the prompt.

\subsection{Full prompt}
Solve a question answering task with interleaving Thought, Interaction with Graph, and Feedback from Graph steps. 

In the Thought step, you should reflect on what further information is needed to answer the question. In the Interaction step, you interact with the graph using one of the following four functions:

\texttt{\{function\_definitions\}}

Here are some examples. You should follow the structure of the examples in your output. 

\texttt{\{demonstrations\}}

\vspace{0.2cm}
Graph Definition: \texttt{\{graph\_definition\}} \\
Question: \texttt{\{question\}} 

\vspace{0.2cm}
Please answer by providing node main features (e.g., names) rather than node IDs. YOU MUST follow the structure of the examples.

\subsection{Graph definitions}
The graph definition describes the structure of each knowledge graph. This includes information like how many types of nodes are in the graph and their associated features.  

\subsubsection{Maple graph description}
There are three types of nodes in the graph: paper, author and venue.\\
Paper nodes have features: title, abstract, year and label. Author nodes have features: name. Venue nodes have features: name.\\
Paper nodes are linked to author nodes, venue nodes, reference nodes and cited\_by nodes. Author nodes are linked to paper nodes. Venue nodes are linked to paper nodes.

\subsubsection{Biomedical graph description}
There are eleven types of nodes in the graph: Anatomy, Biological Process, Cellular Component, Compound, Disease, Gene, Molecular Function, Pathway, Pharmacologic Class, Side Effect, Symptom.\\
Each node has name feature.\\
There are these types of edges: Anatomy-downregulates-Gene, Anatomy-expresses-Gene, Anatomy-upregulates-Gene, Compound-binds-Gene, Compound-causes-Side Effect, Compound-downregulates-Gene, Compound-palliates-Disease, Compound-resembles-Compound, Compound-treats-Disease, Compound-upregulates-Gene, Disease-associates-Gene, Disease-downregulates-Gene, Disease-localizes-Anatomy, Disease-presents-Symptom, Disease-resembles-Disease, Disease-upregulates-Gene, Gene-covaries-Gene, Gene-interacts-Gene, Gene-participates-Biological Process, Gene-participates-Cellular Component, Gene-participates-Molecular Function, Gene-participates-Pathway, Gene-regulates-Gene, Pharmacologic Class-includes-Compound.

\subsubsection{Legal graph description}
There are four types of nodes in the graph: opinion, opinion\_cluster, docket, and court.\\
Opinion nodes have features: plain\_text. Opinion\_cluster nodes have features: syllabus, judges, case\_name, attorneys. Docket nodes have features: pacer\_case\_id, case\_name. Court nodes have features: full\_name, start\_date, end\_date, citation\_string.\\
Opinion nodes are linked to their reference nodes and cited\_by nodes, as well as their opinion\_cluster nodes. Opinion\_cluster nodes are linked to opinion nodes and docket nodes. Docket nodes are linked to opinion\_cluster nodes and court nodes. Court nodes are linked to docket nodes.

\subsubsection{Amazon graph description}
There are two types of nodes in the graph: item and brand.\\
Item nodes have features: title, description, price, img, category. Brand nodes have features: name.\\
Item nodes are linked to their brand nodes, also\_viewed\_item nodes, buy\_after\_viewing\_item nodes, also\_bought\_item nodes, bought\_together\_item nodes. Brand nodes are linked to their item nodes.

\subsubsection{Goodreads graph description}
There are four types of nodes in the graph: book, author, publisher, and series.\\
Book nodes have features: country\_code, language\_code, is\_ebook, title, description, format, num\_pages, publication\_year, url, popular\_shelves, and genres. Author nodes have features: name. Publisher nodes have features: name. Series nodes have features: title and description.\\
Book nodes are linked to their author nodes, publisher nodes, series nodes and similar\_books nodes. Author nodes are linked to their book nodes. Publisher nodes are linked to their book nodes. Series nodes are linked to their book nodes.

\subsubsection{DBLP graph description}
There are three types of nodes in the graph: paper, author and venue.\\
Paper nodes have features: title, abstract, keywords, lang, and year. Author nodes have features: name and organization. Venue nodes have features: name.\\
Paper nodes are linked to their author nodes, venue nodes, reference nodes (the papers this paper cite) and cited\_by nodes (other papers which cite this paper). Author nodes are linked to their paper nodes. Venue nodes are linked to their paper nodes.

\subsection{Function Definitions}

You are a helpful AI bot who can assist in answering questions with help of a knowledge graph. You can answer questions by interacting with the graph using four functions:

(1) RetrieveNode[keyword], which retrieves the related node from the graph according to the corresponding query.

(2) NodeFeature[Node, feature], which returns the detailed attribute information of Node regarding the given "feature" key.

(3) NodeDegree[Node, neighbor\_type], which calculates the number of "neighbor\_type" neighbors of the node Node in the graph.

(4) NeighbourCheck[Node, neighbor\_type], which lists the "neighbor\_type" neighbours of the node Node in the graph and returns them.

You answer these questions with an interleaving Thought, Action taken on the graph using one of the above functions, and an Observation that is returned from the graph. You may take as many steps as necessary to answer the question.

In the Thought step you should come up with a plan to answer the question, this can include what information is still needed or a plan to solving the question. 
In the Action step you should interact with the graph using one of the four functions based on the previous Thought. 
In the Observation step you should provide the information returned from the graph based on the previous Action.

If nothing is returned from the graph, you should try another strategy. e.g. Observation 1: the reference neighbors of 123456 are: []. You should move on to another strategy.

\subsection{Demonstrations}
BEGINNING OF EXAMPLES

\textbf{Example 1} \par
\medskip 

\textbf{Definition of the graph:}
There are three types of nodes in the graph: paper, author and venue. \par
Paper nodes have features: title, abstract, year and label. Author nodes have features: name. Venue nodes have features: name. \par
Paper nodes are linked to their author nodes, venue nodes, reference paper nodes and cited\_by paper nodes. Author nodes are linked to their paper nodes. Venue nodes are linked to their paper nodes. \par

\textbf{Question:}
When was the paper Strongly Interacting Higgs Sector in the Minimal Standard Model published? \par

\textbf{Thought 1:} The question is asking some basic information about a paper. I should first find the paper node related to the paper Strongly Interacting Higgs Sector in the Minimal Standard Model. \par
\textbf{Action 1:} \texttt{RetrieveNode[Strongly Interacting Higgs Sector in the Minimal Standard Model]} \par
\textbf{Observation 1:} The ID of this node is \texttt{3101448248}. \par

\textbf{Thought 2:} The question is asking for the publication year of the paper. I need to check the node feature (year) from the graph. \par
\textbf{Action 2:} \texttt{NodeFeature[3101448248, year]} \par
\textbf{Observation 2:} 1993 \par

\textbf{Thought 3:} The publication year of the paper is 1993. \par
\textbf{Action 3:} \texttt{Finish[1993]} \par

\bigskip 

\textbf{Example 2} \par
\medskip

\textbf{Definition of the graph:}
There are three types of nodes in the graph: paper, author and venue. \par
Paper nodes have features: title, abstract, year and label. Author nodes have features: name. Venue nodes have features: name. \par
Paper nodes are linked to their author nodes, venue nodes, reference paper nodes and cited\_by paper nodes. Author nodes are linked to their paper nodes. Venue nodes are linked to their paper nodes. \par

\textbf{Question:}
How many authors do the paper Mass Accretion Rates in Self-Regulated Disks of T Tauri Stars have? \par

\textbf{Thought 1:} The question is asking for the number of authors of a paper. I should first find the paper node related to the paper Mass Accretion Rates in Self-Regulated Disks of T Tauri Stars in the graph. \par
\textbf{Action 1:} \texttt{RetrieveNode[Mass Accretion Rates in Self-Regulated Disks of T Tauri Stars]} \par
\textbf{Observation 1:} The ID of this node is \texttt{2090642949}. \par

\textbf{Thought 2:} The question is asking for the number of authors of the paper. I should check the node's author neighbor degree from the graph. \par
\textbf{Action 2:} \texttt{NodeDegree[2090642949, author]} \par
\textbf{Observation 2:} 5 \par

\textbf{Thought 3:} The number of authors of the paper is 5. \par
\textbf{Action 3:} \texttt{Finish[5]} \par

\bigskip

\textbf{Example 3} \par
\medskip

\textbf{Definition of the graph:}
There are three types of nodes in the graph: paper, author and venue. \par
Paper nodes have features: title, abstract, year and label. Author nodes have features: name. Venue nodes have features: name. \par
Paper nodes are linked to their author nodes, venue nodes, reference paper nodes and cited\_by paper nodes. Author nodes are linked to their paper nodes. Venue nodes are linked to their paper nodes. \par

\textbf{Question:}
What was the publish venue of the paper Mass Accretion Rates in Self-Regulated Disks of T Tauri Stars? \par

\textbf{Thought 1:} The question is asking information about the node related to the paper Mass Accretion Rates in Self-Regulated Disks of T Tauri Stars. I should first find the paper node related to the paper in the graph. \par
\textbf{Action 1:} \texttt{RetrieveNode[Mass Accretion Rates in Self-Regulated Disks of T Tauri Stars]} \par
\textbf{Observation 1:} The ID of this node is \texttt{2090642949}. \par

\textbf{Thought 2:} The question is asking for the publish venue of the paper. I should check the node's venue neighbor feature from the graph. \par
\textbf{Action 2:} \texttt{NodeFeature[2090642949, venue]} \par
\textbf{Observation 2:} the astrophysical journal, the atmosphere journal \par

\textbf{Thought 3:} The publish venue of the paper is the astrophysical journal, the atmosphere journal. \par
\textbf{Action 3:} \texttt{Finish[the astrophysical journal, the atmosphere journal]} \par

\bigskip

\textbf{Example 4} \par
\medskip

\textbf{Definition of the graph:}
There are three types of nodes in the graph: paper, author and venue. \par
Paper nodes have features: title, abstract, year and label. Author nodes have features: name. Venue nodes have features: name. \par
Paper nodes are linked to their author nodes, venue nodes, reference paper nodes and cited\_by paper nodes. Author nodes are linked to their paper nodes. Venue nodes are linked to their paper nodes. \par

\textbf{Question:}
Who wrote the paper titled ``Strongly Interacting Higgs Sector in the Minimal Standard Model''? \par

\textbf{Thought 1:} The question is asking for the author(s) of a paper. I should first find the paper node related to the paper titled ``Strongly Interacting Higgs Sector in the Minimal Standard Model'' in the graph. \par
\textbf{Action 1:} \texttt{RetrieveNode[Strongly Interacting Higgs Sector in the Minimal Standard Model]} \par
\textbf{Observation 1:} The ID of this node is \texttt{3101448248}. \par

\textbf{Thought 2:} The question is asking for the author(s) of the paper. I should check the node's author neighbor degree from the graph. \par
\textbf{Action 2:} \texttt{NeighbourCheck[3101448248, author]} \par
\textbf{Observation 2:} The author neighbors of \texttt{3101448248} are: [\texttt{'98581309'}, \texttt{'2440536741'}, \texttt{'2345070782'}, \texttt{'2113669516'}, \texttt{'1968129287'}, \texttt{'2112021749'}, \texttt{'2165059783'}, \texttt{'2790497238'}, \texttt{'2035459489'}, \texttt{'2166580174'}]. \par

\textbf{Thought 3:} The author(s) of the paper are: [\texttt{'98581309'}, \texttt{'2440536741'}, \texttt{'2345070782'}, \texttt{'2113669516'}, \texttt{'1968129287'}, \texttt{'2112021749'}]. I should check each of the author nodes to get the author names. \par
\textbf{Action 3:} \texttt{NodeFeature[98581309, name]} \par
\textbf{Observation 3:} Author 1: John F. Gunion \par

\textbf{Thought 4:} The author(s) of the paper are: [\texttt{'98581309'}, \texttt{'2440536741'}, \texttt{'2345070782'}, \texttt{'2113669516'}, \texttt{'1968129287'}, \texttt{'2112021749'}]. I should check each of the author nodes to get the author names. I have already checked the first author. I should check the next author. \par
\textbf{Action 4:} \texttt{NodeFeature[2440536741, name]} \par
\textbf{Observation 4:} Author 2: Howard E. Haber \par

\textbf{Thought 5:} The author(s) of the paper are: [\texttt{'98581309'}, \texttt{'2440536741'}, \texttt{'2345070782'}, \texttt{'2113669516'}, \texttt{'1968129287'}, \texttt{'2112021749'}]. I should check each of the author nodes to get the author names. I have already checked the first two authors. I should check the next author. \par
\textbf{Action 5:} \texttt{NodeFeature[2345070782, name]} \par
\textbf{Observation 5:} Author 3: Gordon Kane \par

\textbf{Thought 6:} The author(s) of the paper are: [\texttt{'98581309'}, \texttt{'2440536741'}, \texttt{'2345070782'}, \texttt{'2113669516'}, \texttt{'1968129287'}, \texttt{'2112021749'}]. I should check each of the author nodes to get the author names. I have already checked the first three authors. I should check the next author. \par
\textbf{Action 6:} \texttt{NodeFeature[2113669516, name]} \par
\textbf{Observation 6:} Author 4: Sally Dawson \par

\textbf{Thought 7:} The author(s) of the paper are: [\texttt{'98581309'}, \texttt{'2440536741'}, \texttt{'2345070782'}, \texttt{'2113669516'}, \texttt{'1968129287'}, \texttt{'2112021749'}]. I should check each of the author nodes to get the author names. I have already checked the first four authors. I should check the next author. \par
\textbf{Action 7:} \texttt{NodeFeature[1968129287, name]} \par
\textbf{Observation 7:} Author 5: Michael S. Chanowitz \par

\textbf{Thought 8:} The author(s) of the paper are: [\texttt{'98581309'}, \texttt{'2440536741'}, \texttt{'2345070782'}, \texttt{'2113669516'}, \texttt{'1968129287'}, \texttt{'2112021749'}]. I should check each of the author nodes to get the author names. I have already checked the first five authors. I should check the next author. \par
\textbf{Action 8:} \texttt{NodeFeature[2112021749, name]} \par
\textbf{Observation 8:} Author 6: Lian-Tao Wang \par

\textbf{Thought 9:} The author(s) of the paper are: [\texttt{'98581309'}, \texttt{'2440536741'}, \texttt{'2345070782'}, \texttt{'2113669516'}, \texttt{'1968129287'}, \texttt{'2112021749'}]. I have checked all the authors and am finished. \par
\textbf{Action 9:} \texttt{Finish[John F. Gunion, Howard E. Haber, Gordon Kane, Sally Dawson, Michael S. Chanowitz, Lian-Tao Wang]} \par

\bigskip

\textbf{Example 5} \par
\medskip

\textbf{Definition of the graph:}
There are three types of nodes in the graph: paper, author and venue. \par
Paper nodes have features: title, abstract, year and label. Author nodes have features: name. Venue nodes have features: name. \par
Paper nodes are linked to their author nodes, venue nodes, reference paper nodes and cited\_by paper nodes. Author nodes are linked to their paper nodes. Venue nodes are linked to their paper nodes. \par

\textbf{Question:}
Considering a reader of the paper Strongly Interacting Higgs Sector in the Minimal Standard Model, Select another paper that the reader would be interested in reading. \par

\textbf{Thought 1:} The question is asking for the papers that a reader of the paper Strongly Interacting Higgs Sector in the Minimal Standard Model would be interested in reading. I should first find the paper node related to the paper Strongly Interacting Higgs Sector in the Minimal Standard Model in the graph. \par
\textbf{Action 1:} \texttt{RetrieveNode[Strongly Interacting Higgs Sector in the Minimal Standard Model]} \par
\textbf{Observation 1:} The ID of this node is \texttt{3101448248}. \par

\textbf{Thought 2:} I should also check the venue of the paper Strongly Interacting Higgs Sector in the Minimal Standard Model to provide the best paper. \par
\textbf{Action 2:} \texttt{NodeFeature[3101448248, venue]} \par
\textbf{Observation 2:} The venue of the paper is Physical Review D. \par

\textbf{Thought 3:} The question is asking for the papers that a reader of the paper would be interested in reading. I should check the node's reference paper neighbors from the graph. \par
\textbf{Action 3:} \texttt{NeighbourCheck[3101448248, reference]} \par
\textbf{Observation 3:} The reference paper neighbors of \texttt{3101448248} are: [\texttt{'1988913734'}, \texttt{'1988913735'}, \texttt{'1988913736'}]. \par

\textbf{Thought 4:} The papers that a reader of the paper would be interested in reading are: [\texttt{'1988913734'}, \texttt{'1988913735'}, \texttt{'1988913736'}]. I should check the first paper. \par
\textbf{Action 4:} \texttt{NodeFeature[1988913734, title]} \par
\textbf{Observation 4:} Paper 1: Advances in Quantum Computing Algorithms \par

\textbf{Thought 5:} The papers that a reader of the paper would be interested in reading are: [\texttt{'1988913735'}, \texttt{'1988913736'}]. I should check the second paper. \par
\textbf{Action 5:} \texttt{NodeFeature[1988913735, title]} \par
\textbf{Observation 5:} Paper 2: Machine Learning Techniques for Big Data Analysis \par

\textbf{Thought 6:} The papers that a reader of the paper would be interested in reading are: [\texttt{'1988913734'}, \texttt{'1988913735'}, \texttt{'1988913736'}]. I should check the third paper. \par
\textbf{Action 6:} \texttt{NodeFeature[1988913736, title]} \par
\textbf{Observation 6:} Paper 3: The Higgs Sector and Hidden Symmetry \par

\textbf{Thought 7:} The venue of the papers Strongly Interacting Higgs Sector in the Minimal Standard Model and The Higgs Sector and Hidden Symmetry are Physical Review D. which is the same. The question asked for one paper so I will select the best paper. \par
\textbf{Action:} \texttt{Finish[The Higgs Sector and Hidden Symmetry]} \par

END OF EXAMPLES

\section{Hyperparameteres}
\label{sec:appendix_hyperparameters}

This section details the hyperparameters and implementation specifics used in our experiments.

For retrieval and indexing, we used MPNet-v2 as the retriever model for semantic retrieval of nodes from the knowledge graphs. Node embeddings were indexed using FAISS \cite{johnson2019billion} for efficient similarity search.

For LLM text generation, the following parameters were consistently applied: a Temperature of 0.7 and a Top-p (nucleus) sampling threshold of 0.9. These settings were chosen to balance diversity and coherence in the generated responses.

As detailed in Section 4.3 and evaluated throughout Section 4, we varied two primary inference scaling parameters. The Number of reasoning steps (Sequential Scaling) typically varied between 10, 25, and 50 steps. The Number of sampled thought-action pairs for majority voting (Parallel Scaling) typically varied between 1 (no parallel scaling/GraphCoT equivalent), 4, 8, and 16 votes. Specific configurations are detailed in Tables 1, 2, and 3.

All experiments were conducted using Python 3.9 and the Huggingface Transformers library (version 4.51.3). All experiments were conducted on NVIDIA A100 and H100 GPUs.

\end{document}